\documentclass[11pt]{article}

\usepackage[final]{acl}

\usepackage{times}
\usepackage{latexsym}

\usepackage[T1]{fontenc}

\usepackage[utf8]{inputenc}

\usepackage{microtype}

\usepackage{inconsolata}

\usepackage{graphicx}
\usepackage{amsfonts}
\usepackage{booktabs}
\usepackage{xcolor}
\usepackage{makecell}

\title{Tracing the complexity profiles of different linguistic phenomena through the intrinsic dimension of LLM representations}

\author{
  \textbf{Marco Baroni\textsuperscript{1,2,*}},
  \textbf{Emily Cheng\textsuperscript{1,*}},
  \textbf{Iria de-Dios-Flores\textsuperscript{1,*}},
  \textbf{Francesca Franzon\textsuperscript{1,*}}
\\
\\
  \textsuperscript{1}Universitat Pompeu Fabra (UPF)\\
  \textsuperscript{2}ICREA
\\
  \textsuperscript{*}Equal contribution
\\
  \small{
    \textbf{Correspondence:} \href{mailto:marco.baroni@upf.edu}{marco.baroni@upf.edu}
  }
}

\begin{document}
\maketitle
\begin{abstract}
We explore intrinsic dimension (ID) of LLM representations as a marker of linguistic complexity. Specifically, we test whether ID differences across model layers reflect well-known complexity contrasts established in (psycho)linguistics: coordination vs.~subordination, right-branching vs.~center-embedding, and unambiguous vs.~ambiguous attachment. Our results on six different LLMs show that these contrasts are consistently reflected in ID differences, with more complex phenomena eliciting higher ID profiles. Notably, ID differences emerge at different points across layers for different contrasts, also reaching their peaks at different stages. 
Further experiments using representational similarity and layer pruning confirm the trends. We conclude that ID is a useful marker of linguistic complexity in LLMs, that it points to similar linguistic processing steps across disparate LLMs, and that it has the potential to differentiate between different types of complexity.
\end{abstract}

\section{Introduction}

Given LLMs' remarkable linguistic skills, there is widespread interest in understanding how language processing unfolds in their inner layers, both because a better understanding of their inner workings could lead to more efficient and controllable models \citep{balestriero2024characterizing,gromov2025the}, and because LLMs are imperfect but practical in silico models of the human faculty of language (see \citealp{Futrell:Mahowald:2025,ziv2026large} for contrasting views on their usefulness as cognitive models).

In this context, the \textit{Intrinsic Dimension} (ID) of LLMs' inner representations has been used to gather insights on their internal processing stages \citep[e.g.,][]{Valeriani:etal:2023,Doimo_Serra_Ansuini_Cazzaniga_2024,Cheng:etal:2025}. When datapoints are mapped to high-dimensional representations (e.g., sentences embedded in an LLM layer space), i) they will typically lie near a manifold that effectively occupies a much smaller number of dimensions than the full space, i.e., the intrinsic dimension is much lower than the so-called ambient dimension; and ii) the more complex the data are for the representational system (in our case, the LLM layer), the higher this intrinsic dimension will be. Consequently, we can use ID as a probe for complexity of linguistic processing across LLM layers.

It has been shown that LLMs have consistent ID profiles across their layers, and that these profiles cue phases of more or less abstract linguistic processing. Earlier studies, however, only looked at average model behaviour in response to generic corpus data \citep{cai2021isotropy,Cheng:etal:2025} or at very coarse distinctions between data types \citep{tulchinskii2023intrinsic,Yin:etal:2024,lee-etal-2025-geometric}. Here, we take a more granular view and use ID to characterize how LLMs handle highly controlled complexity contrasts that have long been acknowledged in the (psycho)linguistic literature. In particular, we study, in English, i) the distinction between sentences with subordinated vs.~coordinated clauses; ii) the contrast between center-embedding structures with long-distance agreement and equivalent sentences with right branching; and, finally, iii) sentences with attachment ambiguities compared to sentences where meaning disambiguates the attachment site. 

We find a remarkable degree of consistency in how different LLMs process data characterized by these phenomena, and that their ID profiles point at different processing strategies for each of them. Our evidence from ID is further supported by experiments tracking representational similarities and pruning effects. Overall, our results suggest that sufficiently powerful systems converge to similar ways to handle language, even when they lack explicit priors for it, and consequently, that LLMs can provide us with new insights on the nature of linguistic complexity that can, in turn, complement human sentence processing research.

\section{Related work}
\label{sec:related-work}

\paragraph{Intrinsic dimension} While naturalistic language data appear high-dimensional, the \emph{manifold hypothesis} posits they actually lie near a low-dimensional manifold \citep{Goodfellow:etal:2016}. The \emph{intrinsic dimension} of the data is then the dimension of this possibly nonlinear manifold, that is, the number of degrees of freedom that explain it under minimal information loss \citep{Campadelli_Casiraghi_Ceruti_Rozza_2015}. The manifold hypothesis holds not only for LLM parameter spaces \citep{Aghajanyan:etal:2021,Zhang:etal:2023}, but also for their activations: no matter the model or dataset, existing work shows that LLM representation manifolds have an ID orders-of-magnitude lower than their ambient dimension \citep{cai2021isotropy,Valeriani:etal:2023,Cheng:etal:2025}.  Similar to our work, \citet{lee-etal-2025-geometric} and \citet{Cheng:etal:2023} show that representational ID over layers correlates to formal or psycholinguistic notions of linguistic complexity, in particular, $n$-gram diversity \citep{lee-etal-2025-geometric} and surprisal and learnability \citep{Cheng:etal:2023}. We build directly on the work by \citet{Cheng:etal:2025} who found, across different LLMs, a characteristic per-layer ID profile marked by an ID peak in intermediate layers. Through probing and downstream tasks, they show that this ID peak coincides with the phase where the model is first able to perform complex linguistic tasks, suggesting that the peak cues a stage of deep linguistic processing.

\paragraph{Linguistic probing of LLMs} 
A rich literature has probed the internal representations of linguistic structures in LLMs, demonstrating that they encode aspects of syntactic knowledge and provide insights into where and how this information is stored \citep{Belinkov:Glass:2019,Linzen:Baroni:2020,Rogers:etal:2020,Ferrando:etal:2024,Li:Subramani:2025,Simon:etal:2025}. %
Early work by~\citet{Hewitt:Manning:2019} introduced a method based on linear probing, showing that bidirectional models represent information about syntactic dependencies and the relation between the involved constituents. 
Further research on more recent models and other probing techniques has revealed a more nuanced picture of syntactic representation, showing that although LLMs do encode syntactic information, they are also sensitive to the interference of local cues like closely occurring words~\citep{agarwal2025mechanisms,Simon:etal:2025}.
However, models can still pick up structural differences in sentences that look similar on the surface, and capture the subtle meaning changes they produce, showing an ability to integrate syntactic and semantic processing~\cite{kennedy2025evidence}.
Minimal-pair probing provides a useful way to track specific aspects of linguistic processing and representations in minimally varying contexts: 
\citet{He:etal:2024} compared how models process minimal pairs contrasting a well-formed sentence with an otherwise identical sentence including a targeted grammatical violation. Through this method, they showed that some information about syntactic competence is represented in early layers, and that as sentences become more complex, the models need more layers to evaluate their grammaticality. Crucially, features at the interface between syntax and semantics are more difficult for models to learn than purely syntactic patterns~\citep{graichen2026grammartransformerssystematicreview}.

\paragraph{Linguistic complexity} 

Linguistic complexity can be examined from two complementary perspectives. On the one hand, it can be understood as the degree of elaboration of a grammatical system at the different levels of representation, often known as formal complexity. On the other hand, it can be defined in terms of the effort required by language users to process or learn a given linguistic structure, often known as functional or processing complexity.  Formal complexity pertains to the structural and computational properties of grammar. This involves, for instance, the number of rules, the degree of hierarchical embedding, the number of idiosyncrasies in the system, or the length and depth of constituent and sentence structure \citep{Culicover2014,Hawkins2014, TrotzkeandZwart2014}. Functional or processing complexity, in turn, concerns how such structures are implemented and experienced by language users, shaping parsing, memory, acquisition, and neural activity \citep{Hawkins2014, MennandDuffield2014}. The two perspectives are interrelated: formal descriptions delimit what structures a grammar makes available, while functional accounts reveal how these structures are deployed and constrained in real-time language processing. 
Given the strong interplay between these dimensions of complexity, we will treat complexity holistically, assuming that a given structure is more complex than its baseline irrespective of the specific type of complexity involved. We nonetheless return to the distinction between formal and functional complexity in the Discussion.

\begin{table*}[!htbp]
\centering
\small
\setlength{\tabcolsep}{6pt} 
\begin{tabular}{@{}lp{11cm}@{}}
\toprule
\textbf{Contrast} & \textbf{Example} \\
\midrule
\textcolor[HTML]{8AB6E6}{Coordination}     
& \textcolor[HTML]{8AB6E6}{(1) The blacksmith is babbling and the politicians are doubting and the tutor is writing and the banker is listening} \\
\textcolor[HTML]{293C7E}{\textbf{Subordination}}    
& \textcolor[HTML]{293C7E}{(2) The blacksmith is babbling that the politicians are doubting that the tutor is writing that the banker is listening} \\
\midrule
\textcolor[HTML]{E7A8E2}{Right branching}  
& \textcolor[HTML]{E7A8E2}{(3) The politicians advised the potters that were waiting} \\
\textcolor[HTML]{A03A9D}{\textbf{Center embedding}} 
& \textcolor[HTML]{A03A9D}{(4) The potters that the politicians advised were waiting} \\
\midrule
\textcolor[HTML]{A8D65E}{Unambiguous}      
& \textcolor[HTML]{A8D65E}{(5) The boy of the mother who gave birth got cold.} \\
\textcolor[HTML]{2F7F3E}{\textbf{Ambiguous}}        
& \textcolor[HTML]{2F7F3E}{(6) The friend of the mother who gave birth got cold.} \\
\bottomrule
\end{tabular}
\caption{Examples of the experimental manipulations in the three main datasets.}
\label{tab:experimental_manipulations}
\end{table*}

\section{Experimental materials}
\label{sec:experimental-materials}

We designed three main datasets instantiating the complexity contrasts shown in Table \ref{tab:experimental_manipulations}.\footnote{Our datasets are available from: \url{https://github.com/franfranz/syntactic_complexity_in_LLMs}}
These contrasts are implemented through classic minimal pairs, allowing us to isolate the contribution of each linguistic manipulation. We report results for these datasets in the main text, and present additional dataset variants and analyses in the appendices.

\paragraph{\textcolor[HTML]{8AB6E6}{\textbf{Coordination}} vs.~\textcolor[HTML]{293C7E}{\textbf{subordination}}} 

These clause-combining operations yield syntactic configurations differing in hierarchical depth. Coordinated sentences are the result of an iterative process generating a flat structure. Subordinated sentences result from a recursive process generating an embedded structure (examples (1) and (2) in Table \ref{tab:experimental_manipulations}). In this sense, subordination corresponds to a higher level of \textit{formal} complexity \citep{TrotzkeandZwart2014}. In spite of the near-total lexical overlap, the hierarchical differences between coordinated and subordinated structures lead to clear meaning differences. For example, since coordination links clauses of equal syntactic status, swapping them leaves meaning essentially unchanged. In contrast, the hierarchical dependency created by embedding one clause within another becomes evident when the clauses are swapped. Our primary datasets consist of sentences  which only vary in the conjunction used (\textit{and} vs.~\textit{that}). Details are in App.~\ref{app:nesting-dataset}. Supplementary manipulations (App.~\ref{app:length_coord_subord}) involve varying the number of clauses that are combined (from 2 to 4) and contrasting \textit{or} (instead of \textit{and}) with \textit{that}. 

\paragraph{\textcolor[HTML]{E7A8E2}{\textbf{Right-branching}}  vs.~\textcolor[HTML]{A03A9D}{\textbf{center-embedding}}} 

These structures (examples (3) and (4) in Table \ref{tab:experimental_manipulations}) differ sharply in their processing-related, or \textit{functional}, complexity, despite being formally similar in terms of hierarchical depth and the type of underlying dependency relations: the two contain a relative clause (RC) modifying one of the NPs in the main clause. Yet, while right-branching sentences extend linearly by placing the RC at the end, the RC in center-embedding sentences appears between the subject and the verb, creating a long-distance dependency. Even though this alternation leaves sentence meaning essentially intact (differing only in information-structural choices about which noun phrase is in focus), a vast amount of psycholinguistic literature has shown that center-embedding sentences are harder to process due to increased integration and memory storage cost \citep[e.g.,][]{Gibson1998, LewisVasishth2005}. Our dataset creation procedure is detailed in App.~\ref{app:agreement-dataset}. A further experiment (App.~\ref{app:src_orc}) instantiates a similar complexity contrast using subject vs.~object relative clauses, with the former representing the less complex condition. 

\paragraph{\textcolor[HTML]{A8D65E}{\textbf{Unambiguous}} vs.~\textcolor[HTML]{2F7F3E}{\textbf{ambiguous}}} 

RC attachment is a common source of ambiguity that arises when a RC can modify more than one preceding NP. This ambiguity can be eliminated when only one NP is semantically compatible with the content of the RC, illustrating how cues at the syntax–semantics interface guide attachment decisions. Our datasets contain minimal pairs differing only in whether the attachment is lexically disambiguated in favor of the closer NP (example (5) in Table \ref{tab:experimental_manipulations}, a case of so-called ``low'' attachment) or whether both NPs remain semantically plausible, thereby sustaining the ambiguity (example (6) in Table \ref{tab:experimental_manipulations}).\footnote{We also include an unambiguous condition favoring attachment to the first NP, so-called ``high'' attachment (App. \ref{app:ambiguity-dataset}). We focus on the comparison between ambiguous and low-attachment sentences because low attachment reflects the typical attachment preference in English, and thus provides a natural baseline. High-attachment sentences behave similarly to low-attachment ones (App. \ref{app:id_ambiguity}).} Although the two sentences are formally parallel, many psycholinguistic studies have shown that semantically unambiguous RCs permit faster attachment, and are thus less complex, whereas ambiguous ones generate processing slow-downs due to increased competition between attachment sites \citep[e.g.,][]{Gibson1996, CarreirasClifton1999}. We thus consider this complexity contrast more \textit{functional} in nature. Dataset details are in App.~\ref{app:ambiguity-dataset}.

\paragraph{Higher complexity elicits higher LLM surprisal} We ran a preliminary check that LLM processing aligned with the predictions from the (psycho)linguistic literature sketched above. Indeed, for all datasets, all LLMs showed a higher per-token surprisal for the more complex condition as determined by a one-sided t-test ($\alpha=0.05$). Exact surprisal values are given in Table \ref{tab:surprisal} of App.~\ref{app:surprisals}.

\section{Methods}
\label{sec:methods}

Given an LLM and a complexity contrast (e.g., coordination vs.~subordination), we examine whether and how the evolution of representation geometry over layers differs between the hard'' and easy'' conditions. To track this evolution, we focus on the residual stream, the main representational pathway that is progressively updated through residual connections. To do so, for each dataset-model pair, we first extract the layerwise last-token embeddings in the residual stream \citep{elhage2021mathematical}.
We use last-token representations as they are the only ones to attend to the entire sequence, and the ones used to predict the next token. For a dataset of size $N$ and an LLM with hidden dimension $D$, this yields a sequence of representations in $\mathbb R^{N\times D}$. We compute the ID and between-contrast representational similarities using these representations.

\paragraph{Intrinsic dimension} In order to compare LLM processing complexity profiles in different conditions, for each model-dataset-layer combination, we compute the ID of the representations. Among the many ID estimators developed \citep{Campadelli_Casiraghi_Ceruti_Rozza_2015}, we choose the TwoNN estimator \citep{Facco2017EstimatingTI}, as it makes minimal assumptions (only local data uniformity up to the second nearest neighbor), correlates highly to other estimators \citep{Cheng:etal:2023}, and has been widely used in past work estimating the ID of data representations \citep{Valeriani:etal:2023,chen2024sudden,lee-etal-2025-geometric}. Details are in App.~\ref{app:twonn}.

\paragraph{Representational similarity} Because we consider data \emph{manifolds}, we are interested in a notion of \emph{local} representational similarity. We make use of the Information Imbalance ($\Delta$) \citep{Glielmo_Zeni_Cheng_Csányi_Laio_2022}, a measure of neighborhood divergence between representation spaces. $\Delta$ is \emph{asymmetric}: in general, $\Delta(A\to B) \neq \Delta(B \to A)$. As such, $\Delta$ can be thought of as a \emph{directional} distance between spaces. Intuitively, it quantifies the amount of information about one space (e.g., the space of coordinated sentences) that is also captured in another space (e.g., the space of matched subordinated sentence). Both \citet{Cheng:etal:2025} and \citet{acevedo2026a} found $\Delta$ to provide a clearer signal than the more widely used Centered Kernel Alignment measure \citep{Kornblith:etal:2019} in the context of comparing LLM representations. Details are in App.~\ref{app:information_imbalance}.

\paragraph{Models} We test six LLMs in sizes ranging from 7 to 14 billion parameters: Gemma-2-9b (Gemma) \citep{gemmateam2024gemma2improvingopen}, Llama-3-8B (Llama) \citep{grattafiori2024llama3herdmodels}, OLMo-2-13B (OLMo) \citep{walsh2025}, Mistral-7B-v0.1 (Mistral) \citep{jiang2023mistral7b}, Pythia-12B (Pythia) \citep{Biderman:etal:2023}, and Qwen-2.5-14B (Qwen) \citep{qwen2025qwen25technicalreport}. These models may differ in specific architectural choices, for instance nonlinear activation or positional embedding, but all consist of decoder layers that each include an attention and feedforward module. Our use of residual stream representations abstracts away from these specific architectural choices to allow high-level comparison of the models' layerwise dynamics.

\section{Experiments}

Our observations are always based on inspecting the behavior of all six studied LLMs. However, for space reasons, we display the results for Llama, OLMo and Pythia in the main text, and Gemma, Mistral and Qwen in appendices.

\subsection{Intrinsic dimension}
\label{sec:id-results}

\begin{figure*}[tb]
   \centering
   \includegraphics[height=0.46\textheight]{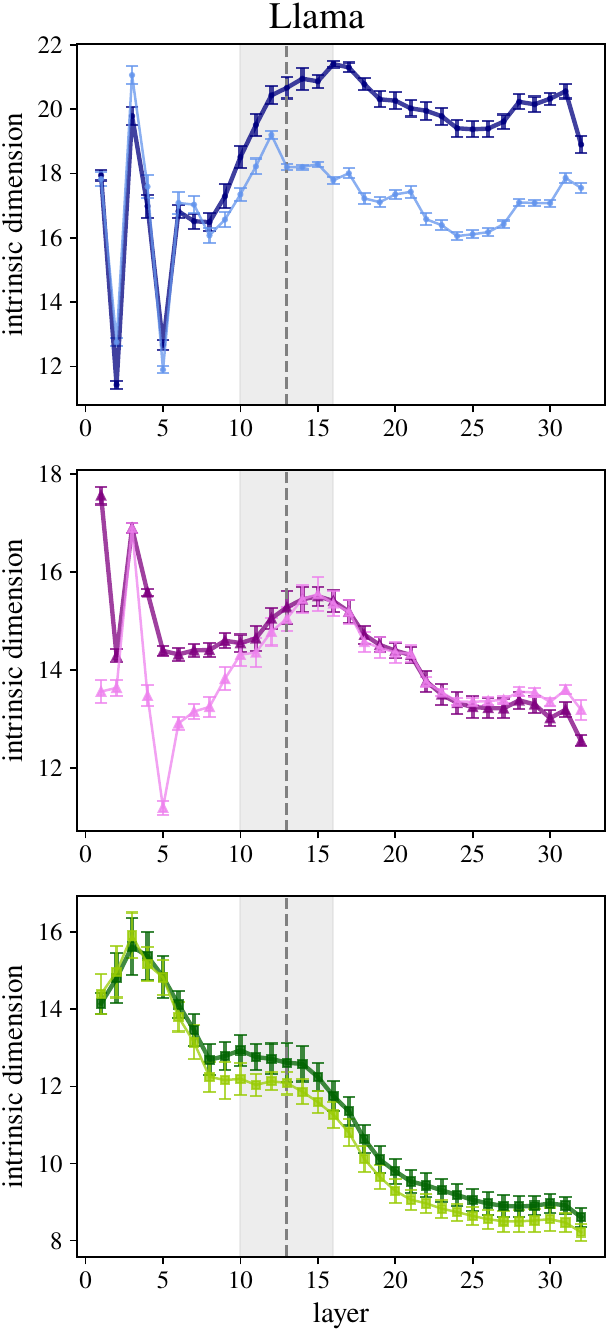}
   \hspace{1pt}
   \includegraphics[height=0.46\textheight]{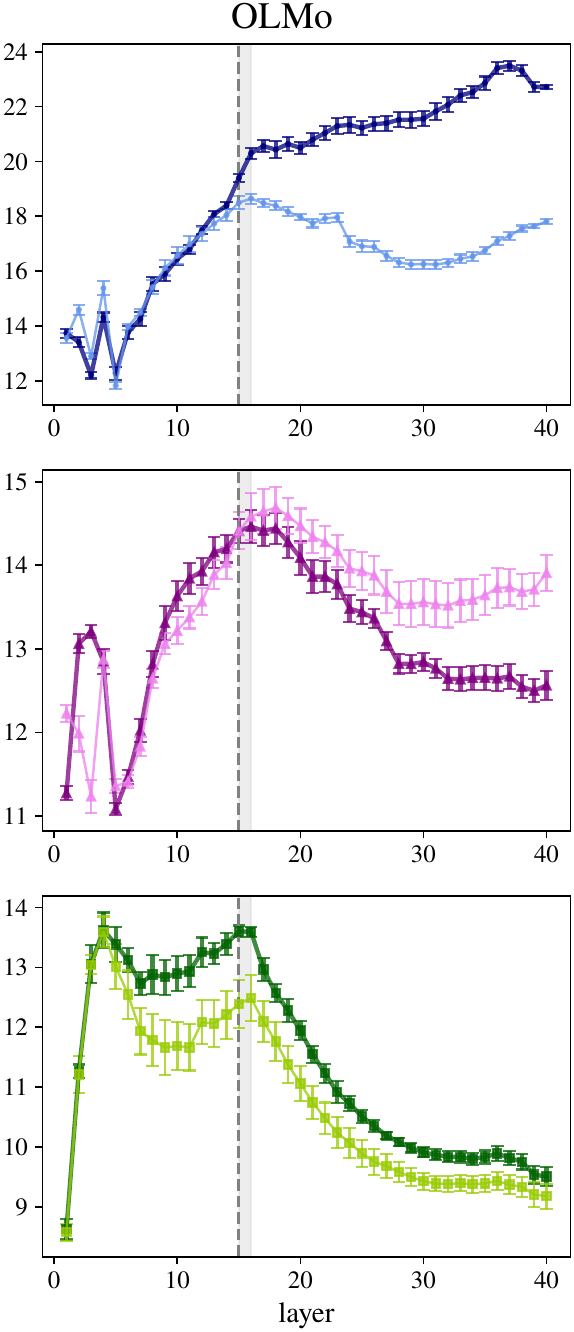}
   \hspace{1pt}
   \includegraphics[height=0.46\textheight]{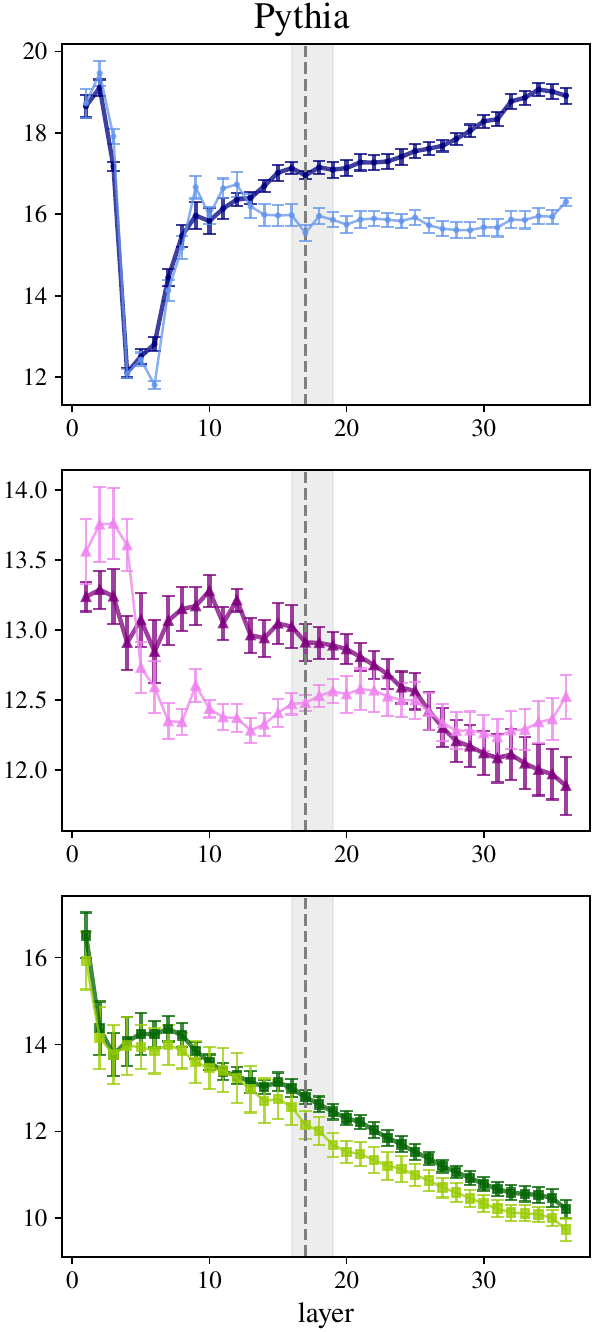}
     \vspace{1mm}
  \includegraphics[width=1.1\textwidth]{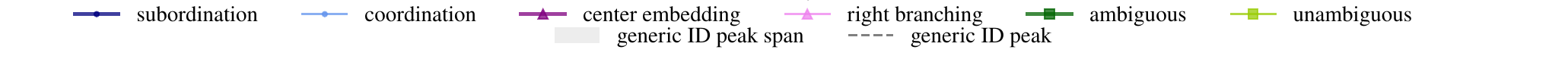}
    \caption{ID profiles through LLM layers (means and error bars across 5 partitions). Vertical dashed line marks maximum ID on generic sequences, and shaded area the corresponding span, estimated as explained in App.~\ref{app:id_peak}.}
    \label{fig:id_profiles_main}
\end{figure*}

\paragraph{Generic sequences} We build on \citet{Cheng:etal:2025}, who showed that, when fed random naturalistic input from a corpus, LLMs systematically display a profile with (at least) one ID ``peak'' that marks an area of deep linguistic processing. We replicate this result in App.~\ref{app:id_peak}, extending it to 3 LLMs that were not studied by Cheng and colleagues. In the same appendix, we confirm their observation that there is a broad alignment between the ID peaks and top performance phases in three semantic and syntactic probing tasks. In all the following experiments, we thus take the generic-sequence ID-peak span across LLM layers (automatically identified as explained in App.~\ref{app:id_peak}) as a reference point for a phase in which the model is focusing on deeper linguistic processing of the inputs, as opposed to surface-level input processing or output generation.
 
Before analyzing our contrast-specific datasets, we note that, in absolute terms, the maximum ID of the various datasets is always orders of magnitudes smaller than the ambient dimension, and it largely correlates with the length of the inputs. It is highest for the 100-word-long generic sequences, next-higher for the 18-word-long 4-clause coordinated/subordinated sentences, and then lower and comparable for the right-branching/center-embedding inputs (8 words) and the unambiguous/ambiguous inputs (13 words on average).\footnote{The ID of the the ambiguity pairs is comparable to that of the slightly shorter right-branching/center-embedding pairs, if not lower, probably because the ambiguity datasets contain less data-points than the others (App.~\ref{app:ambiguity-dataset}). In any case, our analysis focuses on the relative ID \textit{profile} over layers, rather than on absolute ID values.}

\paragraph{\textcolor[HTML]{8AB6E6}{\textbf{Coordination}} vs.~\textcolor[HTML]{293C7E}{\textbf{subordination}}} The first rows of Fig.~\ref{fig:id_profiles_main} and Fig.~\ref{fig:id_profiles_app} (App.~\ref{app:id_profiles}) show how ID evolves for this contrast across layers.\footnote{Figures \ref{fig:length_coord_subord} and  \ref{fig:or_and_that} (App.~\ref{app:length_coord_subord}) show ID profiles for different numbers of clauses being combined, and when clauses are coordinated with \textit{or} instead of \textit{and}, respectively.} After some initial fluctuations possibly due to estimation noise, we observe a very consistent pattern across LLMs: the coordination and subordination ID curves first overlap, but they eventually diverge, typically under or just before the generic-ID peak (considerably earlier for Qwen only). From that point onward, ID is clearly higher for the more complex subordination condition than for coordination.

\paragraph{\textcolor[HTML]{E7A8E2}{\textbf{Right-branching}}  vs.~\textcolor[HTML]{A03A9D}{\textbf{center-embedding}}} This contrast is shown in the middle rows of Fig.~\ref{fig:id_profiles_main} and Fig.~\ref{fig:id_profiles_app} (App.~\ref{app:id_profiles}). After some initial fluctuation, we observe the more complex center-embedding condition to reach a higher ID than the right-branching one. However, this distinction is present only until the curves reach the generic-ID peak (for Pythia, a bit later). After the generic-ID peak, the two curves merge or, for 4/6 models, there is even a \textit{reversal}, whereby right-branching ID is higher. Intriguingly, right-branching sentences end with a nested clause (``The politicians advised the potters [that were waiting]''), unlike the center-embedding ones (``The potters [that the politicians advised] were waiting''). This suggests that the higher right-branching ID after the generic-ID peak might be related to the presence of a nested clause at the end of the sentence, analogously to what we just saw for subordination (recall that we are measuring the ID of the \textit{last} token, that will be inside the nested clause in right-branching sentences only).

\paragraph{\textcolor[HTML]{A8D65E}{\textbf{Unambiguous}} vs.~\textcolor[HTML]{2F7F3E}{\textbf{ambiguous}}} Results are in the last rows of Fig.~\ref{fig:id_profiles_main} and Fig.~\ref{fig:id_profiles_app}. 
We observe a general downward trend of ID, that in most cases becomes clearer after the generic-ID peak. We do not attribute this pattern directly to the attachment manipulation. Unlike in the previous datasets (coordination vs.~subordination and right-branching vs.~center-embedding), where sentence structure is tightly controlled and the final token is always a present participle, the sentences in the present datasets display more variation in the final tokens (e.g., they may be nouns, adverbs, or adjectives: see Table \ref{tab:age-gender-bias-examples}, App.~\ref{app:ambiguity-dataset}). Although such factors are controlled within each minimal pair, allowing us to interpret differences in ID profiles between conditions, they vary across pairs. As a result, they may shape the overall ID trajectory independently of the contrast under investigation.
Notably, the consistently higher ID observed for the ambiguous condition suggests that the models are sensitive to the presence of ambiguity, thus supporting the use of this dataset to probe how LLMs represent it. 

More directly relevant to our investigation, we observe a general tendency for the ID of the more complex ambiguous condition to dominate the simpler unambiguous condition. In this case, however, the generic-ID-peak phase is not acting as a clear delimiter between different ID patterns, and the difference between the curves tends to be small.

\paragraph{}In sum, ID confirms our predictions based on the (psycho)linguistic literature, but in very distinct ways for the three contrasts considered. Syntactic nesting emerges as the linguistic phenomenon most consistently affecting ID profiles: processing deeply nested structure plausibly requires more complex features, leading to higher ID profiles for subordinated structures. Further, the differentiation between flatter and more nested sentences emerges over the layers that coincide with the generic-ID peak, which was proposed as a signature of deeper linguistic processing. It figures that nested structure processing is part of this phase.

ID also responds to long-distance agreement resolution, but in this case the complexity effect is visible earlier, typically before the generic-ID peak. From a strictly structural point of view, center-embedded and right-branching sentences are identical, and their meaning only differs in terms of which element is foregrounded. Thus, the early ID differentiation might be a marker of a lower-level, word-bound kind of processing, linked to matching agreement features between the main subject and its predicate. Intriguingly, for most models we observe higher post-generic-peak ID for right-branching sentences, which we linked to their final nesting, thus relating it to the ID increase  of the subordinated sentences above.

Finally, yet another ID pattern appears for ambiguity resolution. Ambiguous and unambiguous profiles are similar, but the ambiguous constructions have higher ID across the board. This pattern is open to multiple interpretations, but it suggests a contrast that is not linked to specific phases of linguistic processing, but rather to the instability triggered by ambiguity (e.g., the higher complexity could be triggered by keeping both sentence interpretations open until the end).

\subsection{Representation similarity across layers}


\begin{figure*}[tb] 
\centering
\includegraphics[height=0.3\textheight]{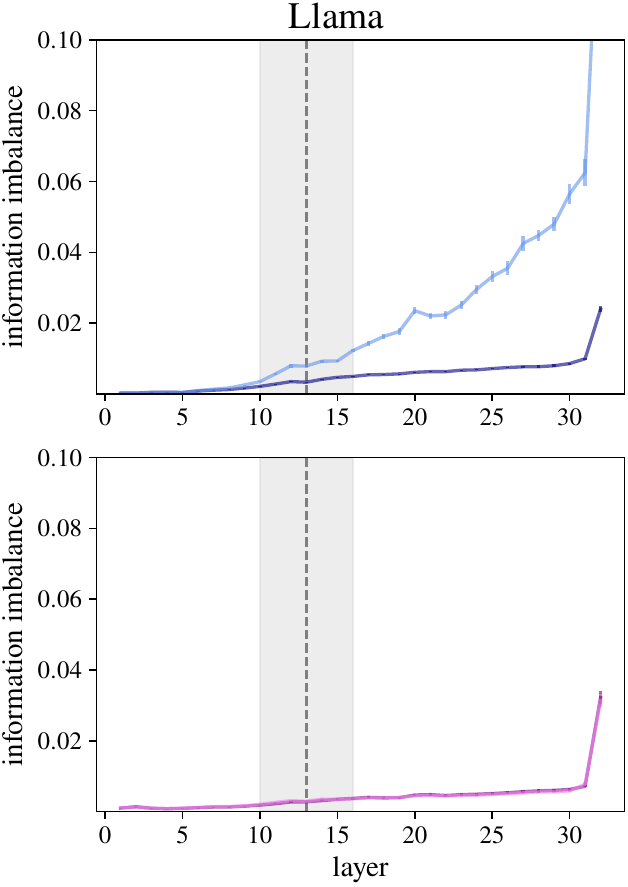}\hspace{2pt}
\includegraphics[height=0.3\textheight]{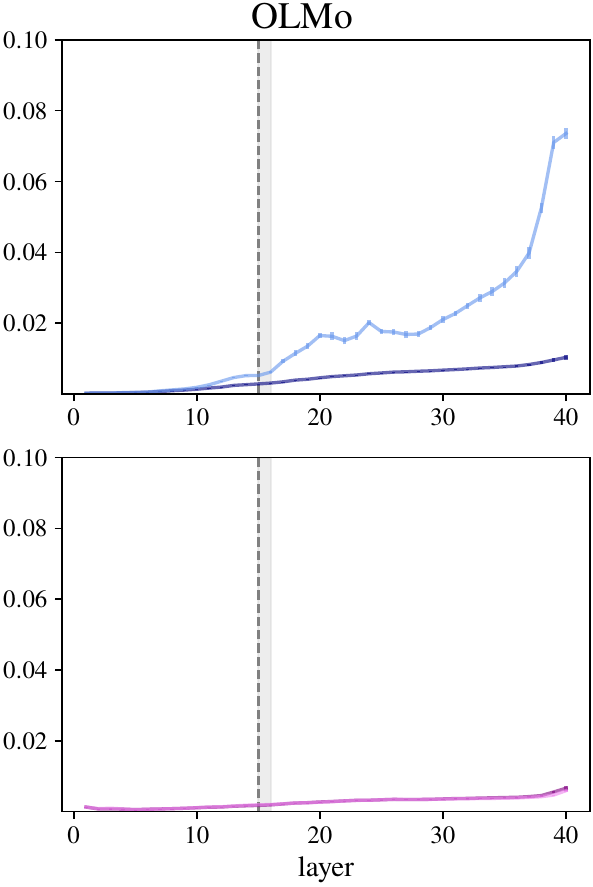}\hspace{2pt}
\includegraphics[height=0.3\textheight]{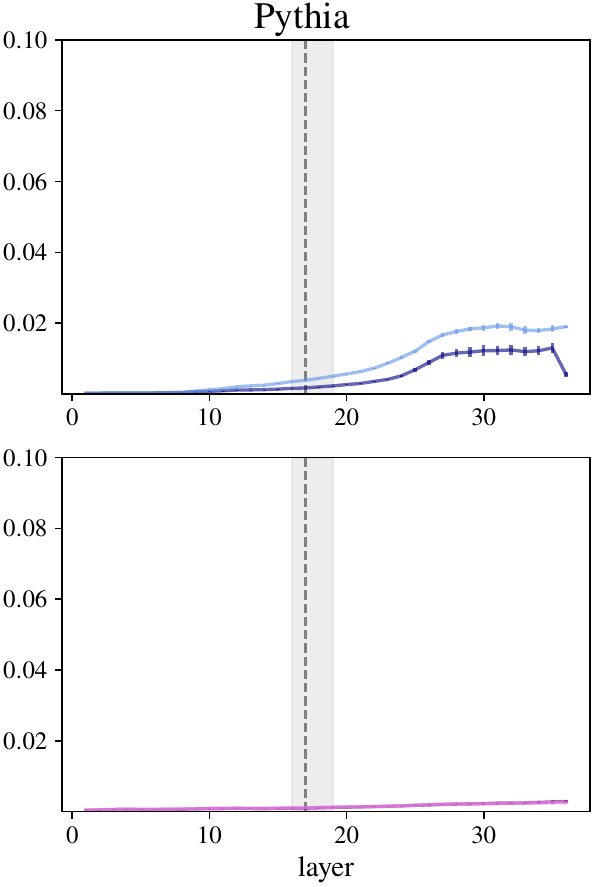}
\vspace{-1mm}
\includegraphics[width=0.88\textwidth]{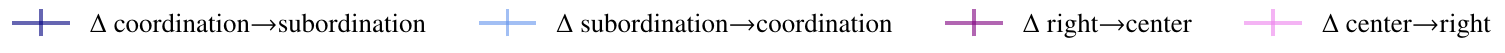}
\includegraphics[width=0.88\textwidth]{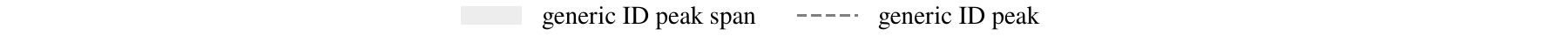}
  
\caption{Information Imbalance $\Delta$ between coordinated/subordinated (top) and right-branching/center-embedding sentences (bottom): means across 5 partitions with error bars (present, but hardly visible). Shaded area marks generic ID-peak span, with a vertical dashed line at the generic-ID maximum. Higher $\Delta$ means lower similarity.}
  \label{fig:ii_profiles_main}
\end{figure*}

\begin{figure*}[tb]
\centering
 \includegraphics[width=\textwidth]{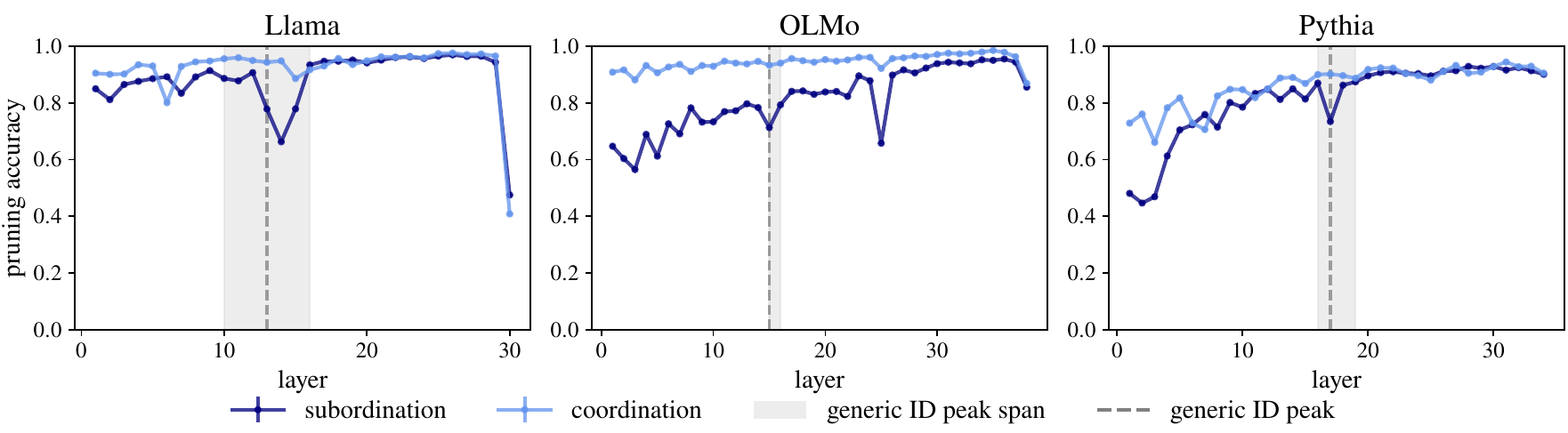}
\caption{Pruning accuracy (proportion of cases in which pruning a layer did not change next-token prediction) of  {\textcolor[HTML]{8AB6E6}{\textbf{coordinated}}/\textcolor[HTML]{293C7E}{\textbf{subordinated}}} sentences. Means and (hardly visible) standard error bars over 5 partitions.}
  \label{fig:ablation_profiles_main}
\end{figure*}

Our claim that ID profiles reflect a genuine difference in LLM processing of different complexity types is supported by a follow-up experiment in which we tracked representational similarity between matched subordinated vs.~coordinated and right-branching vs.~center-embedding sentences.\footnote{The sentences in the ambiguity datasets have significant differences in lexical material, and thus they cannot be naturally paired to measure representational similarity. In App.~\ref{app:ii_validation}, we provide independent empirical validation for the Information Imbalance measure we use here, showing how it approaches 0 for near paraphrases, and 1 for unrelated sentences.}

The top rows of Figures \ref{fig:ii_profiles_main} (and \ref{fig:ii_profiles_app} in App.~\ref{app:ii_profiles}) show the evolution of the Information Imbalance $\Delta$ across layers for the coordinated and subordinated sentence datasets. $\Delta$ is generally extremely low in both directions, i.e., not surprisingly, the model representations of sentence pairs such as those in Table \ref{tab:that-and-examples} (App.~\ref{app:nesting-dataset}), that share word order and virtually all lexical materials, are extremely similar. To put this degree of similarity into perspective, according to the theoretical simulations of \citet{acevedo2026a}, a $\Delta$ of 0.1 (the largest value on our $y$ axes) corresponds to 90\% of shared features. There is also, however, a clear asymmetric divergence in $\Delta$ profiles for 5/6 models. Just under the generic-ID peak, and thus just where the ID difference between the coordination and subordination sets emerges, $\Delta$(\textit{coord}$\to{}$\textit{subord}) becomes progressively \textit{larger} than $\Delta$(\textit{subord}$\to{}$\textit{coord}), except for Qwen (Fig.~\ref{fig:ii_profiles_app}), where the generic-ID peak only marks a phase in which representations diverge, but there is no asymmetry. Thus, in general, the representations of the two sentence types grow apart, and the information contained in the coordinated sentences predict increasingly \textit{less} that of the corresponding subordinated sentences. This makes intuitive sense: if I say that ``\textit{The blacksmith is babbling that the politicians are doubting}'', I am approximately providing the same information as in the corresponding coordinated sentence (``\textit{The blacksmith is babbling and the politicians are doubting}''), but on top of that I am also specifying \textit{what} it is that the blacksmith is babbling, a piece of information that is not contained in the coordinated sentence.

Looking next at the bottom rows of the figures (center-embedding vs.~right-branching), we again observe very strong similarity, decreasing only slightly across the layers. Importantly, in this case there is no asymmetry. This suggests that the LLMs capture the fact that pairs such as ``\textit{The potters that the politicians advised were waiting}'' and ``\textit{The politicians advised the potters that were waiting}'' carry the same denotational meaning.

Finally, in both cases, we observe that $\Delta$ sometimes spikes on the very last layers. This is possibly due to the fact that, as the models must switch to predicting a concrete next token on these layers, and this is likely different for the compared conditions, they must represent different superficial information at the very end.

In sum, at the generic-ID-peak, the LLMs start building structures that make subordinated and coordinated sentences diverge in representation, with the representations of the more richly structured subordinated sentences becoming more predictive of the simpler coordinated sentences than the reverse. On the other hand, no asymmetry is observed for the center-embedding vs.~right-branching contrast, and only a very weak divergence in $\Delta$ across the layers, in line with the strong semantic and structural similarity between the conditions. Again, this suggests a stronger sensitivity of the LLMs to the nesting contrast in their deeper layers.


\subsection{Pruning experiments}

We perform an intervention experiment, looking at the effect of pruning layers that the ID and $\Delta$ tests singled out as crucial for complexity processing. In practice, we remove in turn each layer of an LLM, so that the output from the previous layer is directly fed into the next one. As standard in prior work \citep{gromov2025the,lad2025remarkable,csordas2025do}, we iteratively prune each layer $l$ by removing layer $l$'s attention $+$ MLP contribution, and directly feeding residual stream representations from the previous layer $l-1$ to the next layer $l+1$. We then compute a \textit{pruning accuracy} score by counting the proportion of inputs for which the pruned LLM predicts the same next token as the intact network. Note that, when the pruned-model prediction is different from the original one, there is no guarantee that the latter is better. However, the measure still captures the extent to which pruning changes an LLM behaviour.\footnote{In App.~\ref{app:kl_ablation}, we report similar results obtained with a continuous pruning score, namely KL divergence.}

Fig.~\ref{fig:ablation_profiles_main} (and Fig.~\ref{fig:ablation_coord_subord_appendix} in App.~\ref{app:ablation_profiles}) show pruning profiles for the coordination/subordination conditions. For all models, we observe a tendency for pruning to have a stronger effect on subordinated than coordinated sentences, in line with the other markers of extra complexity for this condition (the effect is typically clearer in the earlier layers). Strikingly, for 4/6 LLMs there is a clear dip in accuracy when pruning layers in the the generic-ID peak phase, bringing further support for the hypothesis that this is a stage in which different LLMs are performing structure-building operations that are especially important for highly nested sentences. Of the two outliers (Fig.~\ref{fig:ablation_coord_subord_appendix} in the appendix), Qwen shows a subordination-specific dip just after the generic-ID peak, whereas in the case of Gemma there is a dip late in the peak, but it is actually stronger for coordination.

Moving to the other contrasts, Figures \ref{fig:ablation_center_right} and \ref{fig:ablation_ambiguity} (App.~\ref{app:ablation_profiles}) do not show  differential pruning effects across conditions (right-branching vs.~center-embedding and unambiguous vs.~ambiguous), and there are no layer-specific dips, except for strong initial- and final-layer effects.

To sum up, we confirm the special status of nesting-driven complexity, and the crucial role that the layers under the generic-ID peak have in determining nested input processing.



\section{Discussion}
\label{sec:discussion}
We showed that different pre-trained LLMs display similar ID profiles, and that the latter are signatures of consistent processing stages. We refined this observation by showing that generic-ID peaks, previously conjectured to be the locus of abstract linguistic processing, correspond to areas where the contrast between flatter (coordinated) and more nested (subordinated) structures emerge. The other contrasts we studied (right branching vs.~center embedding and unambiguous vs.~ambiguous relative clause attachment) also have consistent ID signatures, that however differ and are less clearly related to the generic-ID peak stage. Intriguingly, as discussed in Sec.~\ref{sec:experimental-materials}, the distinction between subordination and the other complexity factors corresponds to a distinction between a more formal type and two more functional ones. Future work should conduct more case studies along this divide, to ascertain whether the ID differences we found are general properties of formal vs.~functional complexity.

From an LLM interpretability perspective, our results shed light on the mechanisms behind the consistent ID profiles that are observed across model layers, and contribute to recent debates on the ``universality'' of LLM representations \citep{Huh:etal:2024}, suggesting that the conjectured universality is at least in part due to similarities in the ways in which LLMs organize linguistic processing. From a linguistic point of view, we have shown that 
ID estimation in LLM representations can provide a new tool to track different degrees of linguistic complexity, a task that is still seen as an open problem in this area \citep{Newmeyer:Preston:2014}.

We believe that, even if LLMs cannot be straightforwardly treated as models of human cognition \citep{ziv2026large}, they still can serve as probes for understanding the structure and complexity of linguistic signals. From this perspective, ID profiles can reveal properties of language representation that are relevant to cognitive science, even if the exact mechanisms giving rise to the representations might be different from those found in humans.


\section*{Limitations} 

Due to compute restrictions, we cannot extend our experiments to larger language models than the ones we studied. However, future work should ascertain the extent to which the observed patterns depend on model size, for families of related LLMs, and track their emergence during training, for LLMs that make their intermediate checkpoints available.

Our results should also be reproduced in languages other than English, and with more naturalistic data than the one we used here. Even in English, we have only started scraping the surface of the set of possible linguistic phenomena that could be studied under the ID lens, and future work should expand the list of case studies to demonstrate that the relation between increases in ID and linguistic complexity is truly general.

Finally, while the pruning experiments provide a proof of concept that there is a causal link between the presence of an ID peak on a layer and its impact on model behavior, the most important aim of future work should be to establish a clear mechanistic explanation of the relation between intrinsic dimension of representations and the linguistic complexity.

\section*{Acknowledgments}
We are thankful to Caleb Mathewos for his fundamental contribution in building the attachment ambiguity datasets. We also thank the ARR reviewers and area chair, Santiago Acevedo, Gemma Boleda, Corentin Kervadec, Iuri Macocco and our colleagues at COLT for advice and feedback. 

MB, EC, and FF received funding from the European Research Council (ERC) under the European Union’s Horizon 2020 research and innovation programme (grant agreement No.~101019291). FF also received funding from the ERC under grant agreement No.~101201497.
All authors received funding from the Catalan government (AGAUR grant SGR 2021 00470).

\bibliography{custom,marco}

@article{
acevedo2026a,
title={A quantitative analysis of semantic information in deep representations of text and images},
author={Santiago Acevedo and Andrea Mascaretti and Riccardo Rende and Mat{\'e}o Mahaut and Marco Baroni and Alessandro Laio},
journal={Transactions on Machine Learning Research},
issn={2835-8856},
year={2026},
url={https://openreview.net/forum?id=sBnaFSIuGR},
note={}
}

@article{agarwal2025mechanisms,
  title={Mechanisms vs. Outcomes: Probing for Syntax Fails to Explain Performance on Targeted Syntactic Evaluations},
  author={Agarwal, Ananth and Jian, Jasper and Manning, Christopher D and Murty, Shikhar},
  journal={arXiv preprint arXiv:2506.16678},
  year={2025}
}

@article{elhage2021mathematical,
   title={A Mathematical Framework for Transformer Circuits},
   author={Elhage, Nelson and Nanda, Neel and Olsson, Catherine and Henighan, Tom and Joseph, Nicholas and Mann, Ben and Askell, Amanda and Bai, Yuntao and Chen, Anna and Conerly, Tom and DasSarma, Nova and Drain, Dawn and Ganguli, Deep and Hatfield-Dodds, Zac and Hernandez, Danny and Jones, Andy and Kernion, Jackson and Lovitt, Liane and Ndousse, Kamal and Amodei, Dario and Brown, Tom and Clark, Jack and Kaplan, Jared and McCandlish, Sam and Olah, Chris},
   year={2021},
   journal={Transformer Circuits Thread},
   note={https://transformer-circuits.pub/2021/framework/index.html}
}

@article{kennedy2025evidence,
  title={Evidence of Hierarchically-Complex Syntactic Structure Within BERT’s Word Representations},
  author={Kennedy, Mary Kathryn},
  journal={Society for Computation in Linguistics},
  volume={8},
  number={1},
  year={2025},
  publisher={University of Massachusetts Amherst Libraries}
}

@inproceedings{
gromov2025the,
title={The Unreasonable Ineffectiveness of the Deeper Layers},
author={Andrey Gromov and Kushal Tirumala and Hassan Shapourian and Paolo Glorioso and Dan Roberts},
booktitle={The Thirteenth International Conference on Learning Representations},
year={2025},
url={https://openreview.net/forum?id=ngmEcEer8a}
}

@inproceedings{
balestriero2024characterizing,
title={Characterizing Large Language Model Geometry Helps Solve Toxicity Detection and Generation},
author={Randall Balestriero and Romain Cosentino and Sarath Shekkizhar},
booktitle={Forty-first International Conference on Machine Learning},
year={2024},
url={https://openreview.net/forum?id=glfcwSsks8}
}

@inproceedings{
walsh2025,
title={2 {OLM}o 2 Furious ({COLM}{\textquoteright}s Version)},
author={Evan Pete Walsh and Luca Soldaini and Dirk Groeneveld and Kyle Lo and Shane Arora and Akshita Bhagia and Yuling Gu and Shengyi Huang and Matt Jordan and Nathan Lambert and Dustin Schwenk and Oyvind Tafjord and Taira Anderson and David Atkinson and Faeze Brahman and Christopher Clark and Pradeep Dasigi and Nouha Dziri and Allyson Ettinger and Michal Guerquin and David Heineman and Hamish Ivison and Pang Wei Koh and Jiacheng Liu and Saumya Malik and William Merrill and Lester James Validad Miranda and Jacob Morrison and Tyler Murray and Crystal Nam and Jake Poznanski and Valentina Pyatkin and Aman Rangapur and Michael Schmitz and Sam Skjonsberg and David Wadden and Christopher Wilhelm and Michael Wilson and Luke Zettlemoyer and Ali Farhadi and Noah A. Smith and Hannaneh Hajishirzi},
booktitle={Second Conference on Language Modeling},
year={2025},
url={https://openreview.net/forum?id=2ezugTT9kU}
}

@misc{grattafiori2024llama3herdmodels,
      title={The Llama 3 Herd of Models}, 
      author={Aaron Grattafiori and Abhimanyu Dubey and Abhinav Jauhri and Abhinav Pandey and Abhishek Kadian and Ahmad Al-Dahle and Aiesha Letman and Akhil Mathur and Alan Schelten and Alex Vaughan and Amy Yang and Angela Fan and Anirudh Goyal and Anthony Hartshorn and Aobo Yang and Archi Mitra and Archie Sravankumar and Artem Korenev and Arthur Hinsvark and Arun Rao and Aston Zhang and Aurelien Rodriguez and Austen Gregerson and Ava Spataru and Baptiste Roziere and Bethany Biron and Binh Tang and Bobbie Chern and Charlotte Caucheteux and Chaya Nayak and Chloe Bi and Chris Marra and Chris McConnell and Christian Keller and Christophe Touret and Chunyang Wu and Corinne Wong and Cristian Canton Ferrer and Cyrus Nikolaidis and Damien Allonsius and Daniel Song and Danielle Pintz and Danny Livshits and Danny Wyatt and David Esiobu and Dhruv Choudhary and Dhruv Mahajan and Diego Garcia-Olano and Diego Perino and Dieuwke Hupkes and Egor Lakomkin and Ehab AlBadawy and Elina Lobanova and Emily Dinan and Eric Michael Smith and Filip Radenovic and Francisco Guzmán and Frank Zhang and Gabriel Synnaeve and Gabrielle Lee and Georgia Lewis Anderson and Govind Thattai and Graeme Nail and Gregoire Mialon and Guan Pang and Guillem Cucurell and Hailey Nguyen and Hannah Korevaar and Hu Xu and Hugo Touvron and Iliyan Zarov and Imanol Arrieta Ibarra and Isabel Kloumann and Ishan Misra and Ivan Evtimov and Jack Zhang and Jade Copet and Jaewon Lee and Jan Geffert and Jana Vranes and Jason Park and Jay Mahadeokar and Jeet Shah and Jelmer van der Linde and Jennifer Billock and Jenny Hong and Jenya Lee and Jeremy Fu and Jianfeng Chi and Jianyu Huang and Jiawen Liu and Jie Wang and Jiecao Yu and Joanna Bitton and Joe Spisak and Jongsoo Park and Joseph Rocca and Joshua Johnstun and Joshua Saxe and Junteng Jia and Kalyan Vasuden Alwala and Karthik Prasad and Kartikeya Upasani and Kate Plawiak and Ke Li and Kenneth Heafield and Kevin Stone and Khalid El-Arini and Krithika Iyer and Kshitiz Malik and Kuenley Chiu and Kunal Bhalla and Kushal Lakhotia and Lauren Rantala-Yeary and Laurens van der Maaten and Lawrence Chen and Liang Tan and Liz Jenkins and Louis Martin and Lovish Madaan and Lubo Malo and Lukas Blecher and Lukas Landzaat and Luke de Oliveira and Madeline Muzzi and Mahesh Pasupuleti and Mannat Singh and Manohar Paluri and Marcin Kardas and Maria Tsimpoukelli and Mathew Oldham and Mathieu Rita and Maya Pavlova and Melanie Kambadur and Mike Lewis and Min Si and Mitesh Kumar Singh and Mona Hassan and Naman Goyal and Narjes Torabi and Nikolay Bashlykov and Nikolay Bogoychev and Niladri Chatterji and Ning Zhang and Olivier Duchenne and Onur Çelebi and Patrick Alrassy and Pengchuan Zhang and Pengwei Li and Petar Vasic and Peter Weng and Prajjwal Bhargava and Pratik Dubal and Praveen Krishnan and Punit Singh Koura and Puxin Xu and Qing He and Qingxiao Dong and Ragavan Srinivasan and Raj Ganapathy and Ramon Calderer and Ricardo Silveira Cabral and Robert Stojnic and Roberta Raileanu and Rohan Maheswari and Rohit Girdhar and Rohit Patel and Romain Sauvestre and Ronnie Polidoro and Roshan Sumbaly and Ross Taylor and Ruan Silva and Rui Hou and Rui Wang and Saghar Hosseini and Sahana Chennabasappa and Sanjay Singh and Sean Bell and Seohyun Sonia Kim and Sergey Edunov and Shaoliang Nie and Sharan Narang and Sharath Raparthy and Sheng Shen and Shengye Wan and Shruti Bhosale and Shun Zhang and Simon Vandenhende and Soumya Batra and Spencer Whitman and Sten Sootla and Stephane Collot and Suchin Gururangan and Sydney Borodinsky and Tamar Herman and Tara Fowler and Tarek Sheasha and Thomas Georgiou and Thomas Scialom and Tobias Speckbacher and Todor Mihaylov and Tong Xiao and Ujjwal Karn and Vedanuj Goswami and Vibhor Gupta and Vignesh Ramanathan and Viktor Kerkez and Vincent Gonguet and Virginie Do and Vish Vogeti and Vítor Albiero and Vladan Petrovic and Weiwei Chu and Wenhan Xiong and Wenyin Fu and Whitney Meers and Xavier Martinet and Xiaodong Wang and Xiaofang Wang and Xiaoqing Ellen Tan and Xide Xia and Xinfeng Xie and Xuchao Jia and Xuewei Wang and Yaelle Goldschlag and Yashesh Gaur and Yasmine Babaei and Yi Wen and Yiwen Song and Yuchen Zhang and Yue Li and Yuning Mao and Zacharie Delpierre Coudert and Zheng Yan and Zhengxing Chen and Zoe Papakipos and Aaditya Singh and Aayushi Srivastava and Abha Jain and Adam Kelsey and Adam Shajnfeld and Adithya Gangidi and Adolfo Victoria and Ahuva Goldstand and Ajay Menon and Ajay Sharma and Alex Boesenberg and Alexei Baevski and Allie Feinstein and Amanda Kallet and Amit Sangani and Amos Teo and Anam Yunus and Andrei Lupu and Andres Alvarado and Andrew Caples and Andrew Gu and Andrew Ho and Andrew Poulton and Andrew Ryan and Ankit Ramchandani and Annie Dong and Annie Franco and Anuj Goyal and Aparajita Saraf and Arkabandhu Chowdhury and Ashley Gabriel and Ashwin Bharambe and Assaf Eisenman and Azadeh Yazdan and Beau James and Ben Maurer and Benjamin Leonhardi and Bernie Huang and Beth Loyd and Beto De Paola and Bhargavi Paranjape and Bing Liu and Bo Wu and Boyu Ni and Braden Hancock and Bram Wasti and Brandon Spence and Brani Stojkovic and Brian Gamido and Britt Montalvo and Carl Parker and Carly Burton and Catalina Mejia and Ce Liu and Changhan Wang and Changkyu Kim and Chao Zhou and Chester Hu and Ching-Hsiang Chu and Chris Cai and Chris Tindal and Christoph Feichtenhofer and Cynthia Gao and Damon Civin and Dana Beaty and Daniel Kreymer and Daniel Li and David Adkins and David Xu and Davide Testuggine and Delia David and Devi Parikh and Diana Liskovich and Didem Foss and Dingkang Wang and Duc Le and Dustin Holland and Edward Dowling and Eissa Jamil and Elaine Montgomery and Eleonora Presani and Emily Hahn and Emily Wood and Eric-Tuan Le and Erik Brinkman and Esteban Arcaute and Evan Dunbar and Evan Smothers and Fei Sun and Felix Kreuk and Feng Tian and Filippos Kokkinos and Firat Ozgenel and Francesco Caggioni and Frank Kanayet and Frank Seide and Gabriela Medina Florez and Gabriella Schwarz and Gada Badeer and Georgia Swee and Gil Halpern and Grant Herman and Grigory Sizov and Guangyi and Zhang and Guna Lakshminarayanan and Hakan Inan and Hamid Shojanazeri and Han Zou and Hannah Wang and Hanwen Zha and Haroun Habeeb and Harrison Rudolph and Helen Suk and Henry Aspegren and Hunter Goldman and Hongyuan Zhan and Ibrahim Damlaj and Igor Molybog and Igor Tufanov and Ilias Leontiadis and Irina-Elena Veliche and Itai Gat and Jake Weissman and James Geboski and James Kohli and Janice Lam and Japhet Asher and Jean-Baptiste Gaya and Jeff Marcus and Jeff Tang and Jennifer Chan and Jenny Zhen and Jeremy Reizenstein and Jeremy Teboul and Jessica Zhong and Jian Jin and Jingyi Yang and Joe Cummings and Jon Carvill and Jon Shepard and Jonathan McPhie and Jonathan Torres and Josh Ginsburg and Junjie Wang and Kai Wu and Kam Hou U and Karan Saxena and Kartikay Khandelwal and Katayoun Zand and Kathy Matosich and Kaushik Veeraraghavan and Kelly Michelena and Keqian Li and Kiran Jagadeesh and Kun Huang and Kunal Chawla and Kyle Huang and Lailin Chen and Lakshya Garg and Lavender A and Leandro Silva and Lee Bell and Lei Zhang and Liangpeng Guo and Licheng Yu and Liron Moshkovich and Luca Wehrstedt and Madian Khabsa and Manav Avalani and Manish Bhatt and Martynas Mankus and Matan Hasson and Matthew Lennie and Matthias Reso and Maxim Groshev and Maxim Naumov and Maya Lathi and Meghan Keneally and Miao Liu and Michael L. Seltzer and Michal Valko and Michelle Restrepo and Mihir Patel and Mik Vyatskov and Mikayel Samvelyan and Mike Clark and Mike Macey and Mike Wang and Miquel Jubert Hermoso and Mo Metanat and Mohammad Rastegari and Munish Bansal and Nandhini Santhanam and Natascha Parks and Natasha White and Navyata Bawa and Nayan Singhal and Nick Egebo and Nicolas Usunier and Nikhil Mehta and Nikolay Pavlovich Laptev and Ning Dong and Norman Cheng and Oleg Chernoguz and Olivia Hart and Omkar Salpekar and Ozlem Kalinli and Parkin Kent and Parth Parekh and Paul Saab and Pavan Balaji and Pedro Rittner and Philip Bontrager and Pierre Roux and Piotr Dollar and Polina Zvyagina and Prashant Ratanchandani and Pritish Yuvraj and Qian Liang and Rachad Alao and Rachel Rodriguez and Rafi Ayub and Raghotham Murthy and Raghu Nayani and Rahul Mitra and Rangaprabhu Parthasarathy and Raymond Li and Rebekkah Hogan and Robin Battey and Rocky Wang and Russ Howes and Ruty Rinott and Sachin Mehta and Sachin Siby and Sai Jayesh Bondu and Samyak Datta and Sara Chugh and Sara Hunt and Sargun Dhillon and Sasha Sidorov and Satadru Pan and Saurabh Mahajan and Saurabh Verma and Seiji Yamamoto and Sharadh Ramaswamy and Shaun Lindsay and Shaun Lindsay and Sheng Feng and Shenghao Lin and Shengxin Cindy Zha and Shishir Patil and Shiva Shankar and Shuqiang Zhang and Shuqiang Zhang and Sinong Wang and Sneha Agarwal and Soji Sajuyigbe and Soumith Chintala and Stephanie Max and Stephen Chen and Steve Kehoe and Steve Satterfield and Sudarshan Govindaprasad and Sumit Gupta and Summer Deng and Sungmin Cho and Sunny Virk and Suraj Subramanian and Sy Choudhury and Sydney Goldman and Tal Remez and Tamar Glaser and Tamara Best and Thilo Koehler and Thomas Robinson and Tianhe Li and Tianjun Zhang and Tim Matthews and Timothy Chou and Tzook Shaked and Varun Vontimitta and Victoria Ajayi and Victoria Montanez and Vijai Mohan and Vinay Satish Kumar and Vishal Mangla and Vlad Ionescu and Vlad Poenaru and Vlad Tiberiu Mihailescu and Vladimir Ivanov and Wei Li and Wenchen Wang and Wenwen Jiang and Wes Bouaziz and Will Constable and Xiaocheng Tang and Xiaojian Wu and Xiaolan Wang and Xilun Wu and Xinbo Gao and Yaniv Kleinman and Yanjun Chen and Ye Hu and Ye Jia and Ye Qi and Yenda Li and Yilin Zhang and Ying Zhang and Yossi Adi and Youngjin Nam and Yu and Wang and Yu Zhao and Yuchen Hao and Yundi Qian and Yunlu Li and Yuzi He and Zach Rait and Zachary DeVito and Zef Rosnbrick and Zhaoduo Wen and Zhenyu Yang and Zhiwei Zhao and Zhiyu Ma},
      year={2024},
      eprint={2407.21783},
      archivePrefix={arXiv},
      primaryClass={cs.AI},
      url={https://arxiv.org/abs/2407.21783}, 
}

@inproceedings{Doimo_Serra_Ansuini_Cazzaniga_2024, title={The Representation Landscape of Few-Shot Learning and Fine-Tuning in Large Language Models}, volume={37}, url={https://proceedings.neurips.cc/paper_files/paper/2024/file/206018a258033def63607fbdf364bd2d-Paper-Conference.pdf}, DOI={10.52202/079017-0576}, booktitle={Advances in Neural Information Processing Systems}, publisher={Curran Associates, Inc.}, author={Doimo, Diego and Serra, Alessandro and Ansuini, Alessio and Cazzaniga, Alberto}, editor={Globerson, A. and Mackey, L. and Belgrave, D. and Fan, A. and Paquet, U. and Tomczak, J. and Zhang, C.}, year={2024}, pages={18122–18165} }

@misc{gemmateam2024gemma2improvingopen,
      title={Gemma 2: Improving Open Language Models at a Practical Size}, 
      author={Morgane Riviere and Shreya Pathak and Pier Giuseppe Sessa and Cassidy Hardin and Surya Bhupatiraju and Léonard Hussenot and Thomas Mesnard and Bobak Shahriari and Alexandre Ramé and Johan Ferret and Peter Liu and Pouya Tafti and Abe Friesen and Michelle Casbon and Sabela Ramos and Ravin Kumar and Charline Le Lan and Sammy Jerome and Anton Tsitsulin and Nino Vieillard and Piotr Stanczyk and Sertan Girgin and Nikola Momchev and Matt Hoffman and Shantanu Thakoor and Jean-Bastien Grill and Behnam Neyshabur and Olivier Bachem and Alanna Walton and Aliaksei Severyn and Alicia Parrish and Aliya Ahmad and Allen Hutchison and Alvin Abdagic and Amanda Carl and Amy Shen and Andy Brock and Andy Coenen and Anthony Laforge and Antonia Paterson and Ben Bastian and Bilal Piot and Bo Wu and Brandon Royal and Charlie Chen and Chintu Kumar and Chris Perry and Chris Welty and Christopher A. Choquette-Choo and Danila Sinopalnikov and David Weinberger and Dimple Vijaykumar and Dominika Rogozińska and Dustin Herbison and Elisa Bandy and Emma Wang and Eric Noland and Erica Moreira and Evan Senter and Evgenii Eltyshev and Francesco Visin and Gabriel Rasskin and Gary Wei and Glenn Cameron and Gus Martins and Hadi Hashemi and Hanna Klimczak-Plucińska and Harleen Batra and Harsh Dhand and Ivan Nardini and Jacinda Mein and Jack Zhou and James Svensson and Jeff Stanway and Jetha Chan and Jin Peng Zhou and Joana Carrasqueira and Joana Iljazi and Jocelyn Becker and Joe Fernandez and Joost van Amersfoort and Josh Gordon and Josh Lipschultz and Josh Newlan and Ju-yeong Ji and Kareem Mohamed and Kartikeya Badola and Kat Black and Katie Millican and Keelin McDonell and Kelvin Nguyen and Kiranbir Sodhia and Kish Greene and Lars Lowe Sjoesund and Lauren Usui and Laurent Sifre and Lena Heuermann and Leticia Lago and Lilly McNealus and Livio Baldini Soares and Logan Kilpatrick and Lucas Dixon and Luciano Martins and Machel Reid and Manvinder Singh and Mark Iverson and Martin Görner and Mat Velloso and Mateo Wirth and Matt Davidow and Matt Miller and Matthew Rahtz and Matthew Watson and Meg Risdal and Mehran Kazemi and Michael Moynihan and Ming Zhang and Minsuk Kahng and Minwoo Park and Mofi Rahman and Mohit Khatwani and Natalie Dao and Nenshad Bardoliwalla and Nesh Devanathan and Neta Dumai and Nilay Chauhan and Oscar Wahltinez and Pankil Botarda and Parker Barnes and Paul Barham and Paul Michel and Pengchong Jin and Petko Georgiev and Phil Culliton and Pradeep Kuppala and Ramona Comanescu and Ramona Merhej and Reena Jana and Reza Ardeshir Rokni and Rishabh Agarwal and Ryan Mullins and Samaneh Saadat and Sara Mc Carthy and Sarah Cogan and Sarah Perrin and Sébastien M. R. Arnold and Sebastian Krause and Shengyang Dai and Shruti Garg and Shruti Sheth and Sue Ronstrom and Susan Chan and Timothy Jordan and Ting Yu and Tom Eccles and Tom Hennigan and Tomas Kocisky and Tulsee Doshi and Vihan Jain and Vikas Yadav and Vilobh Meshram and Vishal Dharmadhikari and Warren Barkley and Wei Wei and Wenming Ye and Woohyun Han and Woosuk Kwon and Xiang Xu and Zhe Shen and Zhitao Gong and Zichuan Wei and Victor Cotruta and Phoebe Kirk and Anand Rao and Minh Giang and Ludovic Peran and Tris Warkentin and Eli Collins and Joelle Barral and Zoubin Ghahramani and Raia Hadsell and D. Sculley and Jeanine Banks and Anca Dragan and Slav Petrov and Oriol Vinyals and Jeff Dean and Demis Hassabis and Koray Kavukcuoglu and Clement Farabet and Elena Buchatskaya and Sebastian Borgeaud and Noah Fiedel and Armand Joulin and Kathleen Kenealy and Robert Dadashi and Alek Andreev},
      year={2024},
      eprint={2408.00118},
      archivePrefix={arXiv},
      primaryClass={cs.CL},
      url={https://arxiv.org/abs/2408.00118}, 
}

@misc{jiang2023mistral7b,
      title={Mistral 7B}, 
      author={Albert Q. Jiang and Alexandre Sablayrolles and Arthur Mensch and Chris Bamford and Devendra Singh Chaplot and Diego de las Casas and Florian Bressand and Gianna Lengyel and Guillaume Lample and Lucile Saulnier and Lélio Renard Lavaud and Marie-Anne Lachaux and Pierre Stock and Teven Le Scao and Thibaut Lavril and Thomas Wang and Timothée Lacroix and William El Sayed},
      year={2023},
      eprint={2310.06825},
      archivePrefix={arXiv},
      primaryClass={cs.CL},
      url={https://arxiv.org/abs/2310.06825}, 
}

@inproceedings{
csordas2025do,
title={Do Language Models Use Their Depth Efficiently?},
author={R{\'o}bert Csord{\'a}s and Christopher D Manning and Christopher Potts},
booktitle={The Thirty-ninth Annual Conference on Neural Information Processing Systems},
year={2025},
url={https://openreview.net/forum?id=Kz6eUL86XP}
}

@inproceedings{levina_bickel,
 author = {Levina, Elizaveta and Bickel, Peter},
 booktitle = {Advances in Neural Information Processing Systems},
 editor = {L. Saul and Y. Weiss and L. Bottou},
 pages = {},
 publisher = {MIT Press},
 title = {Maximum Likelihood Estimation of Intrinsic Dimension},
 url = {https://proceedings.neurips.cc/paper_files/paper/2004/file/74934548253bcab8490ebd74afed7031-Paper.pdf},
 volume = {17},
 year = {2004}
}

@misc{qwen2025qwen25technicalreport,
      title={Qwen2.5 Technical Report}, 
      author={An Yang and Baosong Yang and Beichen Zhang and Binyuan Hui and Bo Zheng and Bowen Yu and Chengyuan Li and Dayiheng Liu and Fei Huang and Haoran Wei and Huan Lin and Jian Yang and Jianhong Tu and Jianwei Zhang and Jianxin Yang and Jiaxi Yang and Jingren Zhou and Junyang Lin and Kai Dang and Keming Lu and Keqin Bao and Kexin Yang and Le Yu and Mei Li and Mingfeng Xue and Pei Zhang and Qin Zhu and Rui Men and Runji Lin and Tianhao Li and Tianyi Tang and Tingyu Xia and Xingzhang Ren and Xuancheng Ren and Yang Fan and Yang Su and Yichang Zhang and Yu Wan and Yuqiong Liu and Zeyu Cui and Zhenru Zhang and Zihan Qiu},
      year={2025},
      eprint={2412.15115},
      archivePrefix={arXiv},
      primaryClass={cs.CL},
      url={https://arxiv.org/abs/2412.15115}, 
}

@inproceedings{
lad2025remarkable,
title={Remarkable Robustness of {LLM}s: Stages of Inference?},
author={Vedang Lad and Jin Hwa Lee and Wes Gurnee and Max Tegmark},
booktitle={The Thirty-ninth Annual Conference on Neural Information Processing Systems},
year={2025},
url={https://openreview.net/forum?id=Wxh5Xz7NpJ}
}

@article{Glielmo_Zeni_Cheng_Csányi_Laio_2022, title={Ranking the information content of distance measures}, volume={1}, url={https://dx.doi.org/10.1093/pnasnexus/pgac039}, DOI={10.1093/pnasnexus/pgac039}, abstractNote={Abstract. Real-world data typically contain a large number of features that are often heterogeneous in nature, relevance, and also units of measure. When a}, number={2}, journal={PNAS Nexus}, publisher={Oxford Academic}, author={Glielmo, Aldo and Zeni, Claudio and Cheng, Bingqing and Csányi, Gábor and Laio, Alessandro}, year={2022}, month=may, language={en} }

@article{Facco2017EstimatingTI,
  title={Estimating the intrinsic dimension of datasets by a minimal neighborhood information},
  author={Elena Facco and Maria d’Errico and Alex Rodriguez and Alessandro Laio},
  journal={Scientific Reports},
  year={2017},
  volume={7},
  url={https://api.semanticscholar.org/CorpusID:3991422}
}

@inproceedings{
chen2024sudden,
title={Sudden Drops in the Loss: Syntax Acquisition, Phase Transitions, and Simplicity Bias in {MLM}s},
author={Angelica Chen and Ravid Shwartz-Ziv and Kyunghyun Cho and Matthew L Leavitt and Naomi Saphra},
booktitle={The Twelfth International Conference on Learning Representations},
year={2024},
url={https://openreview.net/forum?id=MO5PiKHELW}
}

@article{Campadelli_Casiraghi_Ceruti_Rozza_2015, title={Intrinsic Dimension Estimation: Relevant Techniques and a Benchmark Framework}, volume={2015}, ISSN={1024-123X}, DOI={10.1155/2015/759567}, abstractNote={When dealing with datasets comprising high-dimensional points, it is usually advantageous to discover some data structure. A fundamental information needed to this aim is the minimum number of parameters required to describe the data while minimizing the information loss. This number, usually called intrinsic dimension, can be interpreted as the dimension of the manifold from which the input data are supposed to be drawn. Due to its usefulness in many theoretical and practical problems, in the last decades the concept of intrinsic dimension has gained considerable attention in the scientific community, motivating the large number of intrinsic dimensionality estimators proposed in the literature. However, the problem is still open since most techniques cannot efficiently deal with datasets drawn from manifolds of high intrinsic dimension and nonlinearly embedded in higher dimensional spaces. This paper surveys some of the most interesting, widespread used, and advanced state-of-the-art methodologies. Unfortunately, since no benchmark database exists in this research field, an objective comparison among different techniques is not possible. Consequently, we suggest a benchmark framework and apply it to comparatively evaluate relevant state-of-the-art estimators.}, journal={Mathematical Problems in Engineering}, publisher={Hindawi}, author={Campadelli, P. and Casiraghi, E. and Ceruti, C. and Rozza, A.}, year={2015}, month=oct, pages={e759567}, language={en} }

@inproceedings{
cai2021isotropy,
title={Isotropy in the Contextual Embedding Space: Clusters and Manifolds},
author={Xingyu Cai and Jiaji Huang and Yuchen Bian and Kenneth Church},
booktitle={International Conference on Learning Representations},
year={2021},
url={https://openreview.net/forum?id=xYGNO86OWDH}
}

@inproceedings{
tulchinskii2023intrinsic,
title={Intrinsic Dimension Estimation for Robust Detection of {AI}-Generated Texts},
author={Eduard Tulchinskii and Kristian Kuznetsov and Kushnareva Laida and Daniil Cherniavskii and Sergey Nikolenko and Evgeny Burnaev and Serguei Barannikov and Irina Piontkovskaya},
booktitle={Thirty-seventh Conference on Neural Information Processing Systems},
year={2023},
url={https://openreview.net/forum?id=8uOZ0kNji6}
}

@inproceedings{lee-etal-2025-geometric,
    title = "Geometric Signatures of Compositionality Across a Language Model{'}s Lifetime",
    author = "Lee, Jin Hwa  and
      Jiralerspong, Thomas  and
      Yu, Lei  and
      Bengio, Yoshua  and
      Cheng, Emily",
    editor = "Che, Wanxiang  and
      Nabende, Joyce  and
      Shutova, Ekaterina  and
      Pilehvar, Mohammad Taher",
    booktitle = "Proceedings of the 63rd Annual Meeting of the Association for Computational Linguistics (Volume 1: Long Papers)",
    month = jul,
    year = "2025",
    address = "Vienna, Austria",
    publisher = "Association for Computational Linguistics",
    url = "https://aclanthology.org/2025.acl-long.265/",
    doi = "10.18653/v1/2025.acl-long.265",
    pages = "5292--5320",
    ISBN = "979-8-89176-251-0"
}

@incollection{Hawkins2014,
    author = {Hawkins, John A.},
    isbn = {9780199685301},
    title = {Major contributions from formal linguistics to the complexity debate},
    booktitle = {Measuring Grammatical Complexity},
    publisher = {Oxford University Press},
    year = {2014},
    month = {10},
    doi = {10.1093/acprof:oso/9780199685301.003.0002},
    url = {https://doi.org/10.1093/acprof:oso/9780199685301.003.0002},
    eprint = {https://academic.oup.com/book/0/chapter/160438796/chapter-pdf/39533317/acprof-9780199685301-chapter-2.pdf},
}

@incollection{TrotzkeandZwart2014,
    author = {Trotzke, Andreas and Zwart, Jan-Wouter},
    isbn = {9780199685301},
    title = {The complexity of narrow syntax: Minimalism, representational economy, and simplest Merge},
    booktitle = {Measuring Grammatical Complexity},
    publisher = {Oxford University Press},
    year = {2014},
    month = {10},
    doi = {10.1093/acprof:oso/9780199685301.003.0007},
    url = {https://doi.org/10.1093/acprof:oso/9780199685301.003.0007},
    eprint = {https://academic.oup.com/book/0/chapter/160450352/chapter-pdf/39533694/acprof-9780199685301-chapter-7.pdf},
}

@incollection{Culicover2014,
    author = {Culicover, Peter W.},
    isbn = {9780199685301},
    title = {Constructions, complexity, and word order variation},
    booktitle = {Measuring Grammatical Complexity},
    publisher = {Oxford University Press},
    year = {2014},
    month = {10},
    doi = {10.1093/acprof:oso/9780199685301.003.0008},
    url = {https://doi.org/10.1093/acprof:oso/9780199685301.003.0008},
    eprint = {https://academic.oup.com/book/0/chapter/160452253/chapter-pdf/39533709/acprof-9780199685301-chapter-8.pdf},
}

@incollection{MennandDuffield2014,
    author = {Menn, Lise and Duffield, Cecily Jill},
    isbn = {9780199685301},
    title = {Looking for a ‘Gold Standard’ to measure language complexity: what psycholinguistics and neurolinguistics can (and cannot) offer to formal linguistics},
    booktitle = {Measuring Grammatical Complexity},
    publisher = {Oxford University Press},
    year = {2014},
    month = {10},
    doi = {10.1093/acprof:oso/9780199685301.003.0014},
    url = {https://doi.org/10.1093/acprof:oso/9780199685301.003.0014},
    eprint = {https://academic.oup.com/book/0/chapter/160465068/chapter-pdf/39533932/acprof-9780199685301-chapter-14.pdf},
}

@article{Gibson1998,
title = {Linguistic complexity: locality of syntactic dependencies},
journal = {Cognition},
volume = {68},
number = {1},
pages = {1-76},
year = {1998},
issn = {0010-0277},
doi = {https://doi.org/10.1016/S0010-0277(98)00034-1},
url = {https://www.sciencedirect.com/science/article/pii/S0010027798000341},
author = {Edward Gibson}
}

@article{Gibson1996,
  title={Recency preference in the human sentence processing mechanism},
  author={Edward Gibson and Neal J. Pearlmutter and Enriqueta Canseco-Gonzalez and Gregory Hickok},
  journal={Cognition},
  year={1996},
  volume={59},
  pages={23-59},
  url={https://api.semanticscholar.org/CorpusID:45236089}
}

@article{CarreirasClifton1999,   
    title={Another word on parsing relative clauses: Eyetracking evidence from Spanish and English}, 
    volume={27}, 
    DOI={10.3758/bf03198535}, 
    number={5}, 
    journal={Memory and Cognition}, 
    author={Carreiras, Manuel and Clifton, Charles}, 
    year={1999}, 
    month={Sep}, 
    pages={826–833}
}

@ARTICLE{LewisVasishth2005,
  title     = "An activation-based model of sentence processing as skilled
               memory retrieval",
  author    = "Lewis, Richard L and Vasishth, Shravan",
  journal   = "Cogn. Sci.",
  publisher = "Wiley",
  volume    =  29,
  number    =  3,
  pages     = "375--419",
  month     =  may,
  year      =  2005,
  copyright = "http://onlinelibrary.wiley.com/termsAndConditions\#vor",
  language  = "en"
}

@article{lau2021subject,
  title={The subject advantage in relative clauses: A review},
  author={Lau, Elaine and Tanaka, Nozomi},
  journal={Glossa: a journal of general linguistics},
  volume={6},
  number={1},
  year={2021},
  publisher={Open Library of Humanities}
}

@misc{graichen2026grammartransformerssystematicreview,
      title={The Grammar of Transformers: A Systematic Review of Interpretability Research on Syntactic Knowledge in Language Models}, 
      author={Nora Graichen and Iria de-Dios-Flores and Gemma Boleda},
      year={2026},
      eprint={2601.19926},
      archivePrefix={arXiv},
      primaryClass={cs.CL},
      url={https://arxiv.org/abs/2601.19926}, 
}

@article{ziv2026large,
  title   = {Large Language Models as Proxies for Theories of Human Linguistic Cognition},
  author  = {Ziv, Imry and Lan, Nur and Chemla, Emmanuel and Katzir, Roni},
  journal = {Journal of Language Modelling},
  volume  = {14},
  number  = {2},
  year    = {2026},
  doi     = {10.15398/jlm.457}
}

@misc{mathewos2026attachment,
  author       = {Mathewos, Caleb and de-Dios-Flores, Iria},
  title        = {Large Scale Dataset for Attachment Ambiguity},
  year         = {2026},
  publisher    = {Zenodo},
  howpublished = {Zenodo},
  doi          = {10.5281/zenodo.21262989},
  url          = {https://doi.org/10.5281/zenodo.21262989},
}

@inproceedings{Aghajanyan:etal:2021,
    title = "Intrinsic dimensionality explains the effectiveness of language model fine-tuning",
    author = "Aghajanyan, Armen  and
      Gupta, Sonal  and
      Zettlemoyer, Luke",
    booktitle = "Proceedings of ACL",
    year = "2021",
    address = "Online",
    pages = "7319--7328",
}

@InProceedings{Biderman:etal:2023,
  title = 	 {{Pythia}: A suite for analyzing large language models across training and scaling},
  author =       {Biderman, Stella and Schoelkopf, Hailey and Anthony, Quentin Gregory and Bradley, Herbie and O'Brien, Kyle and Hallahan, Eric and Khan, Mohammad Aflah and Purohit, Shivanshu and Prashanth, Usvsn Sai and Raff, Edward and Skowron, Aviya and Sutawika, Lintang and Van Der Wal, Oskar},
  booktitle = 	 {Proceedings of ICML},
  pages = 	 {2397--2430},
  year = 	 {2023},
address = {Honolulu, HI}
}

@article{Belinkov:Glass:2019,
    author = {Belinkov, Yonatan and Glass, James},
    title = {Analysis methods in neural language processing: A survey},
    journal = {Transactions of the Association for Computational Linguistics},
    volume = {7},
    pages = {49--72},
    year = {2019},
}

@inproceedings{Cheng:etal:2023,
    title = "Bridging information-theoretic and geometric compression in language models",
    author = "Cheng, Emily  and
      Kervadec, Corentin  and
      Baroni, Marco",
    booktitle = "Proceedings of EMNLP",
    year = "2023",
    address = "Singapore",
    pages = "12397--12420",
}

@inproceedings{Cheng:etal:2025,
    title = "Emergence of a high-dimensional abstraction phase in language transformers",
    author = "Cheng, Emily  and
              Doimo, Diego and
              Kervadec, Corentin and
              Macocco, Iuri and
              Yu, Jade and
              Laio, Alessandro and
              Baroni,, Marco",
    booktitle = "Proceedings of ICLR",
    year = "2025",
    address = "Singapore",
    note = {Published online: \url{https://openreview.net/group?id=ICLR.cc/2025/Conference}}

}

@InProceedings{Conneau:etal:2018,
  author = 	{Conneau, Alexis
		and Kruszewski, Germ{\'a}n
		and Lample, Guillaume
		and Barrault, Lo{\"i}c
		and Baroni, Marco},
  title = 	"What you can cram into a single {\$}{\&}!{\#}* vector: Probing sentence embeddings for linguistic properties",
  booktitle = 	{Proceedings ACL},
  year = 	{2018},
  pages = 	{2126--2136},
  address = 	{Melbourne, Australia},
}

@misc{Ferrando:etal:2024,
  author    = {Javier Ferrando and Gabriele Sarti and Arianna Bisazza and Marta Costa-juss\'{a}},
  title     =  {A primer on the inner workings of transformer-based language models},
  year      = {2024},
  howpublished={\url{https://arxiv.org/abs/2405.00208}},
}

@misc{Futrell:Mahowald:2025,
  title={How linguistics learned to stop worrying and love the language models},
  author={Richard Futrell and Kyle Mahowald},
  year = 2025,
  howpublished={\url{https://arxiv.org/abs/2501.17047}}
}

@book{Goodfellow:etal:2016,
    title={Deep Learning},
    author={Ian Goodfellow and Yoshua Bengio and Aaron Courville},
    publisher={MIT Press},
    year={2016},
    address    = {Cambridge, MA},
}

@inproceedings{He:etal:2024,
    title = "Decoding probing: Revealing internal linguistic structures in neural language models using minimal pairs",
    author = "He, Linyang  and
      Chen, Peili  and
      Nie, Ercong  and
      Li, Yuanning  and
      Brennan, Jonathan",
    booktitle = "Proceedings of LREC-COLING",
    year = "2024",
    address = "Torino, Italy",
    pages = "4488--4497",
}

@inproceedings{Hewitt:Manning:2019,
    title = "{A} structural probe for finding syntax in word representations",
    author = "Hewitt, John  and
      Manning, Christopher",
    booktitle = "Proceedings of NAACL",
    month = jun,
    year = "2019",
    address = "Minneapolis, MN",
    pages = "4129--4138",
}

@InProceedings{Huh:etal:2024,
  title = 	 {The {Platonic} Representation Hypothesis},
  author =       {Huh, Minyoung and Cheung, Brian and Wang, Tongzhou and Isola, Phillip},
  booktitle = 	 {Proceedings of ICML},
  pages = 	 {20617--20642},
  year = 	 {2024},
  address = {Vienna, Austria},
}

@inproceedings{Kornblith:etal:2019,
  title = 	 {Similarity of Neural Network Representations Revisited},
  author = 	 {Simon Kornblith and Mohammad Norouzi and Honglak Lee and Geoffrey Hinton },
  booktitle = 	 {Proceedings of ICML},
  pages = 	 {3519--3529},
  year = 	 {2019},
  address = {Long Beach, CA},
}

@article{Kullback:Leibler:1951,
  title={On information and sufficiency},
  author={Kullback, Solomon and Leibler, Richard},
  journal={The Annals of Mathematical Statistics},
  volume={22},
  number={1},
  pages={79--86},
  year={1951},
}

@misc{Li:Subramani:2025,
  author    = {Michael Li and Nishant Subramani},
  title     =  {Echoes of {BERT}: Do modern language models rediscover the classical {NLP} pipeline?},
  year      = {2025},
  howpublished={\url{https://arxiv.org/abs/2506.02132}},
}

@article{Linzen:Baroni:2020,
  author = 	 {Tal Linzen and Marco Baroni},
  title = 	 {Syntactic structure from {Deep Learning}},
  journal = 	 {Annual Review of Linguistics},
  year = 	 {2021},
  pages = {195--212},
  volume = 7,
}

@misc{Macocco:etal:2025,
  title={Outlier dimensions favor frequent tokens in language models},
  author={Iuri Macocco and Nora Graichen and Gemma Boleda and Marco Baroni},
  year = 2025,
  howpublished={\url{https://arxiv.org/abs/2503.21718}},
}

@inproceedings{Merity:etal:2016,
  title={Pointer sentinel mixture models},
  author={Stephen Merity and Caiming Xiong and James Bradbury and Richard Socher},
  booktitle={Proceedings of ICLR Conference Track},
  year={2017},
  address={Toulon, France},
  note = {Published online: \url{https://openreview.net/group?id=ICLR.cc/2017/conference}}
}

@book{Newmeyer:Preston:2014,
  editor =	 {Frederick Newmeyer and Laurel Preston},
  title = 	 {Measuring Grammatical Complexity},
  publisher  = 	 {Oxford University Press},
  address={Oxford, UK},
  year = 	 2014
 }

@article{Rogers:etal:2020,
  title={A primer in {BERTology}: {What} we know about how {BERT} works},
  author={Anna Rogers and Olga Kovaleva and Anna Rumshisky},
  journal = "Transactions of the Association for Computational Linguistics",
  volume = "8",
  year = "2020",
  pages = "842--866",
}

@inproceedings{Simon:etal:2025,
title={Probing syntax in large language models: Successes and remaining challenges},
author={Pablo Simon and Emmanuel Chemla and Jean-Remi King and Yair Lakretz},
booktitle={Proceedings of COLM},
address={Montreal, Canada},
year={2025},
note = {Published online \url{https://openreview.net/forum?id=nrZysNmJ0n}}
}

@inproceedings{Valeriani:etal:2023,
 author = {Valeriani, Lucrezia and Doimo, Diego and Cuturello, Francesca and Laio, Alessandro and Ansuini, Alessio and Cazzaniga, Alberto},
 booktitle = {Proceedings of NeurIPS},
 pages = {51234--51252},
 title = {The geometry of hidden representations of large transformer models},
 address = {New Orleans, LA},
 year = {2023}
}

@InProceedings{Yin:etal:2024,
  title = 	 {Characterizing truthfulness in large language model generations with local intrinsic dimension},
  author =       {Yin, Fan and Srinivasa, Jayanth and Chang, Kai-Wei},
  booktitle = 	 {Proceedings of ICML},
  pages = 	 {57069--57084},
  year = 	 {2024},
  address = {Vienna, Austria}
}

@inproceedings{Zhang:etal:2023,
    title = "Fine-tuning happens in tiny subspaces: Exploring intrinsic task-specific subspaces of pre-trained language models",
    author = "Zhang, Zhong  and
      Liu, Bang  and
      Shao, Junming",
    booktitle = "Proceedings of ACL",
    year = "2023",
    address = "Toronto, Canada",
    pages = "1701--1713",
}

\appendix

\section{Datasets}

\subsection{\textcolor[HTML]{8AB6E6}{\textbf{Coordination}} vs.~\textcolor[HTML]{293C7E}{\textbf{subordination}}}
\label{app:nesting-dataset}

Using OpenAI's GPT-4 followed by manual editing, we first built: i) a list of 17 propositional verbs that could also be used as intransitives, such as \textit{babbling} and \textit{dreaming}; ii) a list of 65 pure intransitive verbs taking human subjects (such as \textit{shivering} and \textit{sleeping}); and iii) a list of 74 nouns, divided between 30 proper nouns (\textit{Michael, Sarah}) and 44 profession names (\textit{gardener, programmer}). We randomly sampled elements from these lists to generate 50k 4-clause subordinated and coordinated sentence pairs that matched the following template:

\begin{verbatim}
NP1 PROPVERB1 CONJ NP2 PROPVERB2 CONJ 
NP3 PROPVERB3 CONJ NP4 INTVERB
\end{verbatim}

\noindent{}where the NPs are either proper names or noun phrases formed by \textit{the} followed by a profession noun (randomly in singular or plural form); the PROPVERBs are propositional verbs in the present continuous (in singular or plural form in agreement with the subject); the INTVERB is a pure-intransitive verb, also in the present continuous and agreeing with its subject; and CONJ is either \textit{that} or \textit{and/or}. The words are sampled so that no noun or verb lemma is repeated. In order to maintain uniqueness when removing the clauses in the middle (see below), we also ensured that the tuples formed by $<$NP1, PROPVERB1, NP4, INTVERB$>$ and  $<$NP1, PROPVERB1, NP2, PROPVERB2$>$ were unique. Five random examples from the dataset constructed in this way are shown in Table \ref{tab:that-and-examples}

\begin{table*}
 \centering
   \begin{tabular}{p{\textwidth}}
     \hline
     Quinn is rejoicing and/or/that the surgeon is doubting and/or/that Mary is screaming and/or/that the driver is faltering\\
     The doctors are muttering and/or/that the firefighter is babbling and/or/that Bill is complaining and/or/that the consultants are hesitating\\
     The engineers are singing and/or/that Jordan is dreaming and/or/that the tutors are rejoicing and/or/that the soldier is sliding\\
     The artist is remembering and/or/that Matthew is doubting and/or/that the judge is writing and/or/that Taylor is trembling\\
     Emily is complaining and/or/that Casey is mumbling and/or/that the manager is writing and/or/that the blacksmith is shivering\\
     \hline
   \end{tabular}
   \caption{Examples from the 4-clause \textcolor[HTML]{8AB6E6}{\textbf{coordination}}/\textcolor[HTML]{293C7E}{\textbf{subordination}}   datasets. The 3-clause sentence derived from the first example in this table is: ``Quinn is rejoicing and/that the surgeon is doubting and/that the driver is faltering''. The 2-clause sentence from the same example is: ``Quinn is rejoicing and/that the driver is faltering''.}
   \label{tab:that-and-examples}
\end{table*}

In order to build the shorter 3- and 2-clause sentences analyzed in App.~\ref{app:length_coord_subord}, we simply removed one or two clauses, respectively, from the middle of the 4-clause sentences. We split the datasests into partitions of 10k sentences, and repeat all experiments 5 times, always reporting means and standard errors across the partitions.

\subsection{\textcolor[HTML]{E7A8E2}{\textbf{Right-branching}}  vs.~\textcolor[HTML]{A03A9D}{\textbf{center-embedding}}} 
\label{app:agreement-dataset}

We used the same nouns and intransitive verbs as for the coordination/subordination datasets (see App.~\ref{app:nesting-dataset}). By again querying ChatGPT 4 and manually editing its outputs, we created a list of 100 transitive verbs that take human subjects and objects (\textit{betray}, \textit{fascinate}, \textit{scold}\ldots). We constructed a set of 50K matched center-embedding and right-branching sentences by randomly sampling from these lists according to the following templates:

\begin{verbatim}
CENTER EMBEDDING:
NP1 that NP2 TRVERB INTVERB
RIGHT BRANCHING:
NP2 TRVERB NP1 that INTVERB
\end{verbatim}

\noindent{}where NP1 is formed by \textit{the} followed by a profession noun in plural or singular form; NP2 is either formed in the same way or it is a proper noun; TRVERB is a transitive verb in past-simple form; and INTVERB is a past continuous form of an intransitive verb agreeing in number with NP1. Again, no sentence contains a repeated noun or verb. Using the same data, we constructed an additional contrast between subject relative clauses (SRC) and object relative clauses (ORC). This contrast was derived directly from the center-embedding items. The ORC condition corresponds exactly to the center-embedding dataset, relabeled here to make the contrast in terms of relativization type explicit, rather than the branching configuration. The SRC sentences were created by promoting the relative clause subject to subject position in the main clause, while preserving the predicate and lexical content. Five random examples for each of the two contrasts are given in Table \ref{tab:center-right-examples}.

We split the datasests into partitions of 10k sentences, and repeat all experiments 5 times, always reporting means and standard errors across the partitions.

\begin{table*}
 \centering
   \begin{tabular}{p{0.5\textwidth} p{0.5\textwidth}}

   \multicolumn{1}{c}{\textit{Right-branching}} & \multicolumn{1}{c}{\textit{Center-embedding}}\\
   \hline
Sarah intimidated the potters that were frowning        & The potters that Sarah intimidated were frowning        \\
James harassed the veterinarians that were sulking      & The veterinarians that James harassed were sulking      \\
Bill excluded the driver that was escaping              & The driver that Bill excluded was escaping              \\
Elizabeth praised the foresters that were chuckling     & The foresters that Elizabeth praised were chuckling     \\
The gardeners applauded the blacksmith that was hurrying& The blacksmith that the gardeners applauded was hurrying\\
   \hline

   \multicolumn{1}{c}{\textit{Subject relative clause}} & \multicolumn{1}{c}{\textit{Object relative clause}}\\
   \hline
The politicians that were waiting advised the potters       & The potters that the politicians advised were waiting   \\    
The programmer that was gasping rejected the pharmacist     & The pharmacist that the programmer rejected was gasping  \\
The blacksmith that were relaxing punished the doctors      & The doctors that the blacksmith punished were relaxing   \\
The astronomer that were pacing appointed the blacksmiths   & The blacksmiths that the astronomer appointed were pacing \\
The mechanics that was persisting invited the politician    & The politician that the mechanics invited was persisting  \\
   \hline

   \end{tabular}
   \caption{Examples from the \textcolor[HTML]{E7A8E2}{right-branching}/\textcolor[HTML]{A03A9D}{\textbf{center-embedding}} and the \textcolor[HTML]{E413B3}{subject relative clauses}/\textcolor[HTML]{770177}{\textbf{object relative clauses}} datasets.}
   \label{tab:center-right-examples}
\end{table*}

\subsection{\textcolor[HTML]{A8D65E}{\textbf{Unambiguous}} vs.~\textcolor[HTML]{2F7F3E}{\textbf{ambiguous}}} 
\label{app:ambiguity-dataset}

This dataset contains three attachment conditions: an ambiguous condition, in which both noun phrases are equally plausible relative-clause (RC) antecedents, and two unambiguous conditions, in which semantic biases strongly favor attachment to either NP1 (high attachment) or NP2 (low attachment). All sentences conform to the following template: 
\begin{verbatim}
NP1 of NP2 who RC CONTINUATION
\end{verbatim}

In the main text, we focus on the contrast between ambiguous and low-attachment sentences because English favors low attachment, and, thus, this condition displays the expected and least demanding interpretation, making it a natural baseline. Results including the high-attachment dataset are in App~\ref{app:id_ambiguity}.

Using ChatGPT 4, we generated a list of NPs, relative clauses, and continuations that would encode four types of semantic bias: age, gender, role, and logical contradiction. Bias-specific elements were generated using controlled prompting and subsequently filtered, ranked, and manually validated to ensure semantic clarity, grammaticality, and compatibility across conditions. To construct the final dataset, validated elements were recombined under strict constraints that prevent excessive lexical overlap across sentences, while preserving bias consistency. This procedure generated a total of 10,880 attachment triplets (32,640 sentences), where each triplet shares the same relative clause and continuation. Examples for gender- and age-based semantic disambiguations are given in Table \ref{tab:age-gender-bias-examples} and further details can be found in \citet{mathewos2026attachment}.

\begin{table*}
\centering
\begin{tabular}{p{0.48\textwidth} p{0.48\textwidth}}
\multicolumn{1}{c}{\textit{Age bias}} & \multicolumn{1}{c}{\textit{Gender bias}} \\
\hline
\textbf{Ambiguous}: The neighbor of the grandpa who paid a mortgage stood nearby. 
& \textbf{Ambiguous}: The sister of the heiress who was menstruating cooked rice. \\

\textbf{Low attachment}: The child of the comrade who paid a mortgage stood nearby. 
& \textbf{Low attachment}: The uncle of the maiden who was menstruating cooked rice.\\

\textbf{High attachment}: The uncle of the child who paid a mortgage stood nearby. 
& \textbf{High attachment}: The maiden of the uncle who was menstruating cooked rice. \\
\hline
\end{tabular}
\caption{Examples from relative-clause \textcolor[HTML]{A8D65E}{unambiguous}/\textcolor[HTML]{2F7F3E}{\textbf{ambiguous}} attachment datasets. Low attachment is the unambiguous condition used in the main-text experiments.}
\label{tab:age-gender-bias-examples}
\end{table*}

The sentences were split into 5 equal partitions of 2176 items each. We repeat all experiments 5 times, always reporting means and standard errors across the partitions.

\section{Methods}

\subsection{Code and compute estimates}



\paragraph{Code} To compute intrinsic dimensionality and information imbalance, we used the DadaPy toolkit. Representation extraction was accomplished by trivially adapting the code made available by \citet{Cheng:etal:2025}. Surprisal was computed using the Python \texttt{surprisal} package. In all cases, code was run with default parameters.

URLs and licenses of the used assets are provided in the following list:

\begin{description}
    \item[Gemma] \url{https://huggingface.co/google/gemma-2-9b}; license: gemma
    \item[Llama] \url{https://huggingface.co/meta-llama/Meta-Llama-3-8B}; license: llama3
    \item[Mistral] \url{https://huggingface.co/mistralai/Mistral-7B-v0.1}; license: apache-2.0
    \item[OLMo] \url{https://huggingface.co/allenai/OLMo-2-1124-13Bhttps}; license: apache-2.0 
    \item[Pythia] \url{https://huggingface.co/EleutherAI/pythia-12b-deduped}; license: apache-2.0
    \item[Qwen] \url{https://huggingface.co/Qwen/Qwen2.5-14B}; apache-2.0
    \item[Cheng et al.'s code] \url{ https://github.com/chengemily1/id-llm-abstraction}; license: MIT
    \item[DadaPy] \url{https://github.com/sissa-data-science/DADApy}; license: apache-2.0
    \item[Probing tasks] \url{https://github.com/facebookresearch/SentEval/tree/main/data/probing}; license: bsd
    \item[surprisal] \url{https://github.com/aalok-sathe/surprisal}; license: MIT
\end{description}

\paragraph{Compute} Representation extraction took a few wall-clock hours for each LLM, on 1 or 2 NVIDIA A30 GPUs. The most time-consuming extraction step was to run the pruning experiments, as different sets of representations had to be extracted after removing one layer at a time from each of the models, resulting in about two days of running per model. Probing experiments took between 1 and 2 days per task, running on CPUs. Compute time for the other experiments was negligible.

\subsection{TwoNN ID estimation}
\label{app:twonn}
The TwoNN estimator belongs to a class of so-called ``geometric" ID estimation methods \citep{Campadelli_Casiraghi_Ceruti_Rozza_2015}. This class, which also includes, a.o., the Maximum Likelihood Estimator of \citet{levina_bickel}, to which TwoNN highly correlates in practice \citep{Cheng:etal:2023}, generally works as follows. First, under distributional or regularity assumptions of the space, properties of points on the manifold behave according to some theoretical distribution. The analytic expression of this distribution depends on the ID. Then, the ID can be estimated via maximum likelihood estimation from data.

TwoNN realizes these steps as follows. Assume points on the underlying manifold are distributed as a locally homogeneous Poisson point process, where ``locally" means up to the second nearest neighbor of each point. Let $\delta_k^{(i)}$ be the Euclidean distance between the point $x_i$ and its $k$th nearest neighbor. Then, the distance ratios $\mu_i := \delta_2^{(i)}/\delta_2^{(i)} \in [1,\infty)$ have the cumulative distribution function $F(\mu)=(1-\mu)^d \mathbf 1[\mu \geq 1]$ \citep{Facco2017EstimatingTI}. The ID, given by $d$, is estimated as $d = -\log (1-F(\mu))/\log \mu$ via maximum likelihood estimation over all data points. 

\subsection{Information Imbalance}
\label{app:information_imbalance}

Consider two representation spaces $A$ and $B$ consisting respectively of paired data points $\{x_i\}_{i=1}^N$ and $\{y_i\}_{i=1}^N$. Let $r_{ij}^X$ refer to the neighbor-rank of point $j$ to point $i$ in space $X$. For instance, if $x_j$ is $x_i$'s first nearest neighbor in space $X$, then $r_{ij}^X=1$. Then, $\Delta(A\to B)$ is given by \citep{Glielmo_Zeni_Cheng_Csányi_Laio_2022}:
\begin{equation}
    \Delta(A\to B) := \frac{2}{N^2}\sum_{i=1}^N \sum_{j=1}^N r_{ij}^B\mathbf 1[r_{ij}^A=1]. 
\end{equation}
If nearest neighbors in $A$ are also nearest neighbors in $B$, then $\Delta(A\to B) \approx 0$. If nearest neighbors in $A$ have a uniformly-distributed neighbor rank in $B$, i.e., neighbors in $A$ are uninformative about neighbors in $B$, and $\Delta(A\to B)$ is near $1$. 

\section{Additional results}

\subsection{Surprisal}
\label{app:surprisals}

For each dataset, we sampled 1,000 sentences at random to estimate the mean per-token surprisal under each LLM. We compute the mean surprisal per sentence, then average over all 1,000 sentences to obtain one surprisal value per-sentence. For each model-dataset combination, the distribution of surprisals over the 1,000 sentences was approximately normal, according to a Shapiro-Wilk test with a conservative $p$-value cutoff of $\alpha=0.1$. We then performed a one-sided difference-of-means t-test between the \textbf{more} and \textcolor{gray}{\textbf{less}} complex conditions, finding in all cases that the LLM has higher surprisal on the \textbf{more} complex condition, significant at $\alpha=0.05$. Results are displayed in Table \ref{tab:surprisal}.

\begin{table*}[tb]
\centering
\begin{tabular}{lccccccc}
\hline
Model & \multicolumn{2}{c}{subordination/coordination} & \multicolumn{2}{c}{center embed/right branch} & \multicolumn{3}{c}{ambiguity} \\
 & \textcolor[HTML]{8AB6E6}{coord.} & \textcolor[HTML]{293C7E}{\textbf{subord.}} & \textcolor[HTML]{E7A8E2}{right} & \textcolor[HTML]{A03A9D}{\textbf{center}} & \textcolor[HTML]{A8D65E}{low} & \textcolor[HTML]{A8D65E}{high} & \textcolor[HTML]{2F7F3E}{\textbf{ambiguous}} \\ \hline
 Llama  & $5.39_{\color{gray}0.01}$ & $\underline{5.65}_{\color{gray}0.01}$ & $7.74_{\color{gray}0.02}$ & $\underline{7.98}_{\color{gray}0.02}$ & $6.07_{\color{gray}0.02}$ & $6.08_{\color{gray}0.02}$ & $\underline{6.11}_{\color{gray}0.02}$ \\
OLMo   & $4.87_{\color{gray}0.01}$ & $\underline{5.11}_{\color{gray}0.02}$ & $6.51_{\color{gray}0.02}$ & $\underline{6.75}_{\color{gray}0.02}$ & $5.10_{\color{gray}0.02}$ & $5.16_{\color{gray}0.02}$ & $\underline{5.18}_{\color{gray}0.02}$ \\
Pythia & $4.77_{\color{gray}0.01}$ & $\underline{5.14}_{\color{gray}0.01}$ & $6.66_{\color{gray}0.02}$ & $\underline{7.06}_{\color{gray}0.02}$ & $5.37_{\color{gray}0.02}$ & $5.39_{\color{gray}0.02}$ & $\underline{5.39}_{\color{gray}0.02}$ \\
Gemma  & $6.86_{\color{gray}0.02}$ & $\underline{7.62}_{\color{gray}0.02}$ & $10.35_{\color{gray}0.04}$ & $\underline{10.58}_{\color{gray}0.04}$ & $6.79_{\color{gray}0.03}$ & $6.75_{\color{gray}0.03}$ & $\underline{6.90}_{\color{gray}0.03}$ \\
Mistral& $4.87_{\color{gray}0.01}$ & $\underline{5.22}_{\color{gray}0.02}$ & $6.86_{\color{gray}0.02}$ & $\underline{7.00}_{\color{gray}0.02}$ & $5.50_{\color{gray}0.02}$ & $5.52_{\color{gray}0.02}$ & $\underline{5.58}_{\color{gray}0.02}$ \\
Qwen   & $4.77_{\color{gray}0.01}$ & $\underline{5.05}_{\color{gray}0.01}$ & $6.44_{\color{gray}0.02}$ & $\underline{6.61}_{\color{gray}0.02}$ & $5.13_{\color{gray}0.02}$ & $5.15_{\color{gray}0.02}$ & $\underline{5.17}_{\color{gray}0.02}$ \\
\hline
\end{tabular}
\caption{Average surprisal of contrast datasets. The mean surprisal per-token (nats) $\pm$1 SE is shown for the six LLM rows, for each linguistic phenomenon column and contrast condition subcolumns. Each value is computed from a 1,000-sentence sample, drawn randomly from each contrast condition. For each dataset and model, the highest surprisal value across conditions is underlined. The tested LLMs consistently match expectations from the literature, where \textbf{the more (psycho)linguistically complex condition has higher mean surprisal in every case} and statistical significance was determined by a one-sided t-test ($\alpha=0.05$). In the \textbf{subordination/coordination} datasets, the \emph{subordination} condition, given by a more nested syntactic structure, has higher surprisal than the \emph{coordination} condition. For the \textbf{center/right branching} datasets, the \emph{center embedding} condition has higher surprisal than the \emph{right branching} case. Finally, for the \textbf{ambiguity} datasets, the \emph{ambiguous attachment} condition has higher surprisal under an LLM than the unambiguous high and low attachment cases.}
\label{tab:surprisal}
\end{table*}

\subsection{Intrinsic dimension of generic sequences}
\label{app:id_peak}
\begin{figure*}[t]
    \includegraphics[width=\textwidth]{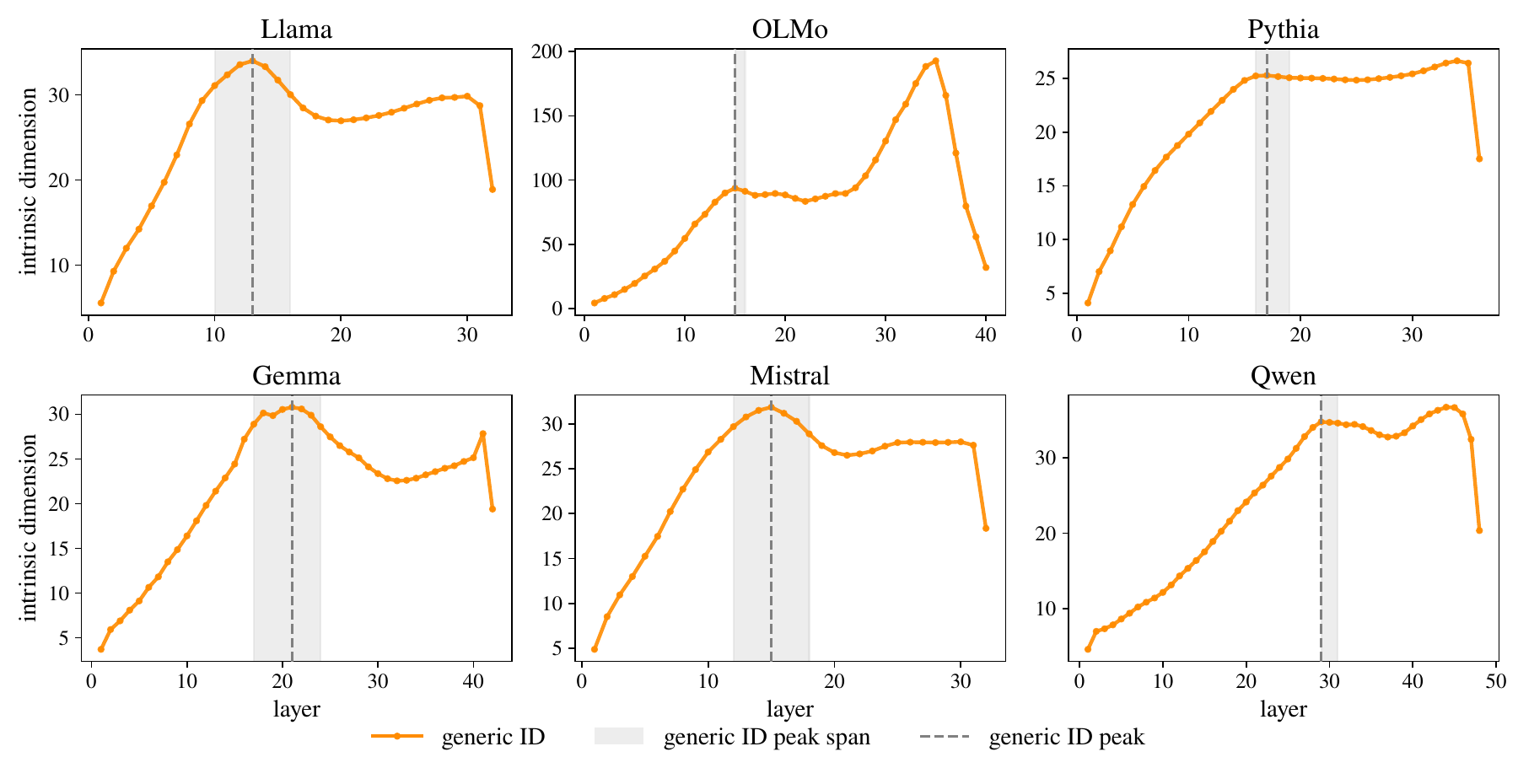}
    \caption{ID values across layers and estimated peak spans for all LLMs given an input of naturalistic corpus sequences. Mean values with (present but virtually invisible) error bars.}
    \label{fig:generic_id}
\end{figure*}

\paragraph{ID profiles} We reproduce the finding by \citet{Valeriani:etal:2023} and \citet{Cheng:etal:2025} that, on generic in-distribution data, a peak in the intrinsic dimension of LLM activations emerges in the intermediate layers. To do so, we use a sample of 50k sequences from the Wikitext corpus \citep{Merity:etal:2016} repurposed from \citet{Macocco:etal:2025}.  Each sequence consists of 100 words and begins with a start of sentence but is not constrained to end with a period, so that the final word can have any part of speech. Then, for each model, on five random non-overlapping data splits of 10k sequences each, we compute the ID using the TwoNN estimator on the last-token representation. This produces an ID profile across layers for each model, shown in Fig.~\ref{fig:generic_id}. Finally, for each model, following \citet{Cheng:etal:2025}, we heuristically demarcate the first ID peak-span (gray in the figures). We locate the nearest inflection points around the maximum using second-order finite differences. Note that, with respect to Cheng and colleagues, we extend these experiments to Gemma, Qwen and a newer and larger version of OLMo, thus further confirming the generality of their observation.

Fig.~\ref{fig:generic_id} first of all confirms that the intrinsic dimension is always orders of magnitude smaller than the ambient ones (which is always $>3.5K$). Importantly, all models have an intrinsic dimensionality peak in their mid layers, followed in some cases (OLMo, Pythia, Qwen) by another late-layer peak, a phenomenon also reported by Cheng and colleagues.

\begin{figure*}[t]
    \includegraphics[width=\textwidth]{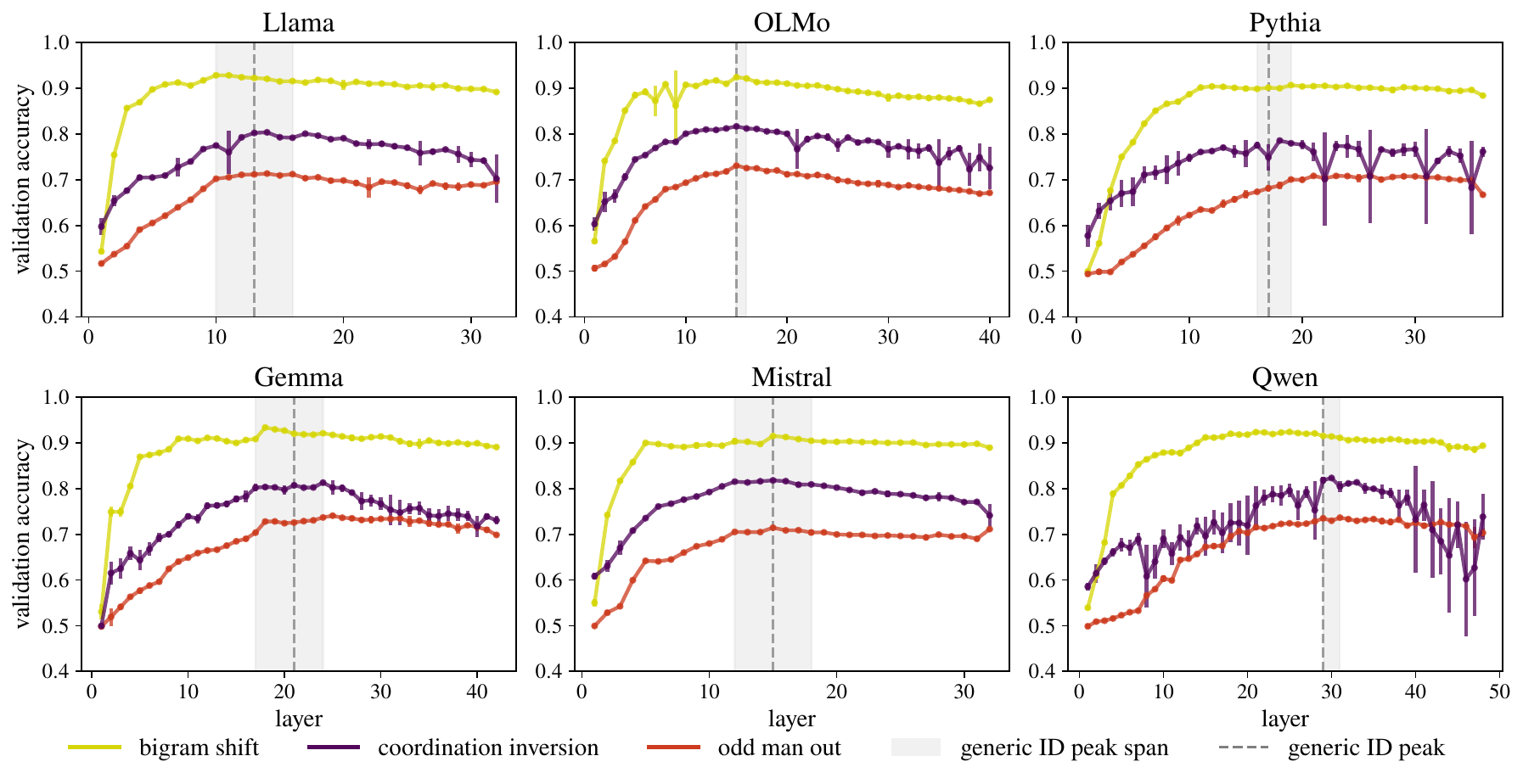}
    \caption{Accuracies in three probing tasks (bigram shift, coordination inversion and odd man out) across layers, for all models. Means and standard errors across five seeds.}
    \label{fig:probing}
\end{figure*}

\paragraph{Higher-order linguistic processing around the ID peak} \Citet{Cheng:etal:2025} found that the ID peak span coincides with a phase in which higher-order linguistic information is made available, as indicated by a set of MLP-based probing tasks and downstream tests. We replicate their probing results for all tested models, using syntactic and semantic tasks from \citet{Conneau:etal:2018} (bigram shift, coordination inversion, and odd man out). While our results (in Fig.~\ref{fig:probing}) are somewhat noisy, we confirm that performance on these tasks tends to grow to its maximum within the ID peak span. This confirms that the generic-sequence ID peak span provides a useful rule-of-thumb for the locus of deeper syntactic/semantic processing in the LLMs.

\subsection{Intrinsic dimension profiles of the other models}
\label{app:id_profiles}

\begin{figure*}[t]
  \centering
    \includegraphics[height=0.46\textheight]{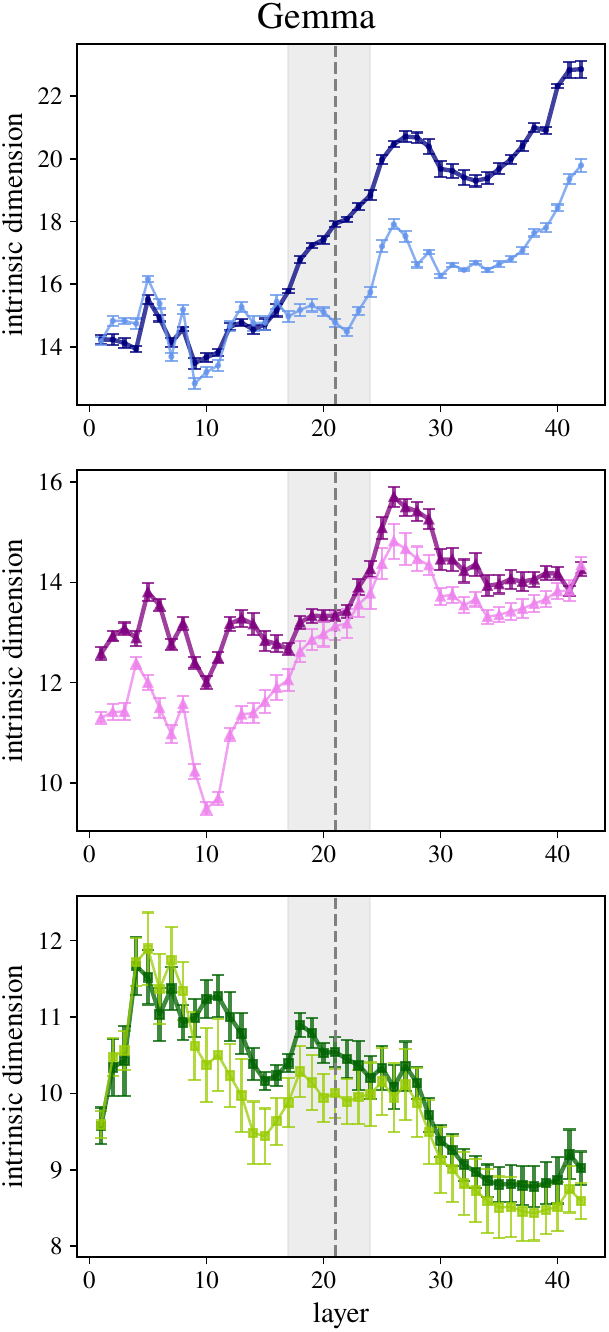}\hspace{1pt}
  \includegraphics[height=0.46\textheight]{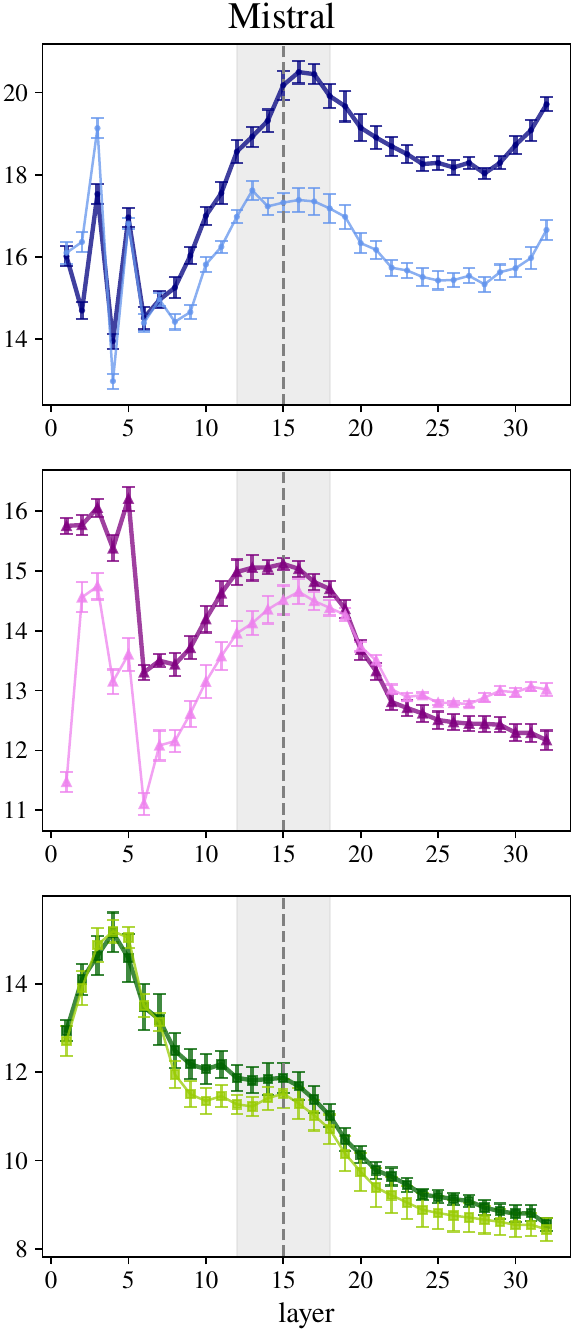}\hspace{1pt}
  \includegraphics[height=0.46\textheight]{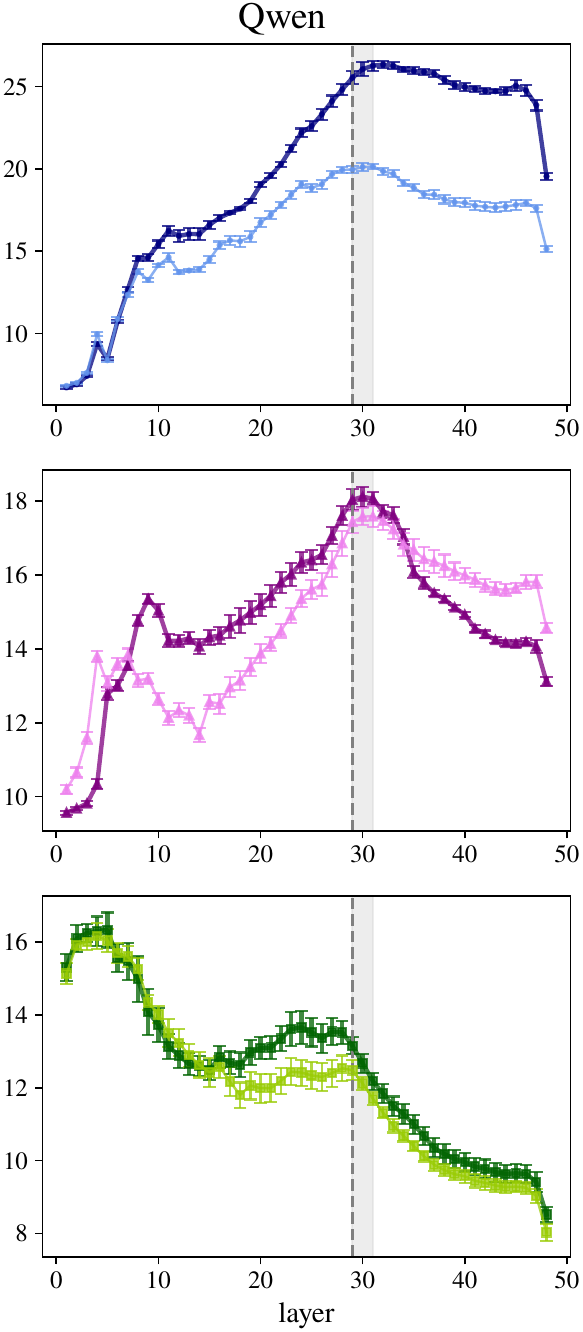}
  \vspace{1mm}
  \includegraphics[width=1.1\textwidth]{figures_v4/legend_id.pdf}
    \caption{ID profiles through LLM layers (means and error bars across 5 partitions). Vertical dashed line marks maximum ID on generic sequences, and shaded area the corresponding span, estimated as explained in App.~\ref{app:id_peak}.}
  \label{fig:id_profiles_app}
\end{figure*}

ID profiles for the 3 models not shown in the main text (Gemma, Mistral and Qwen) are reported in Fig.~\ref{fig:id_profiles_app}.

\subsection{Intrinsic dimension profiles for different levels of embedding and for \textit{or}-coordination}
\label{app:length_coord_subord}

\begin{figure*}[t]
    \centering
    \includegraphics[width = \textwidth]{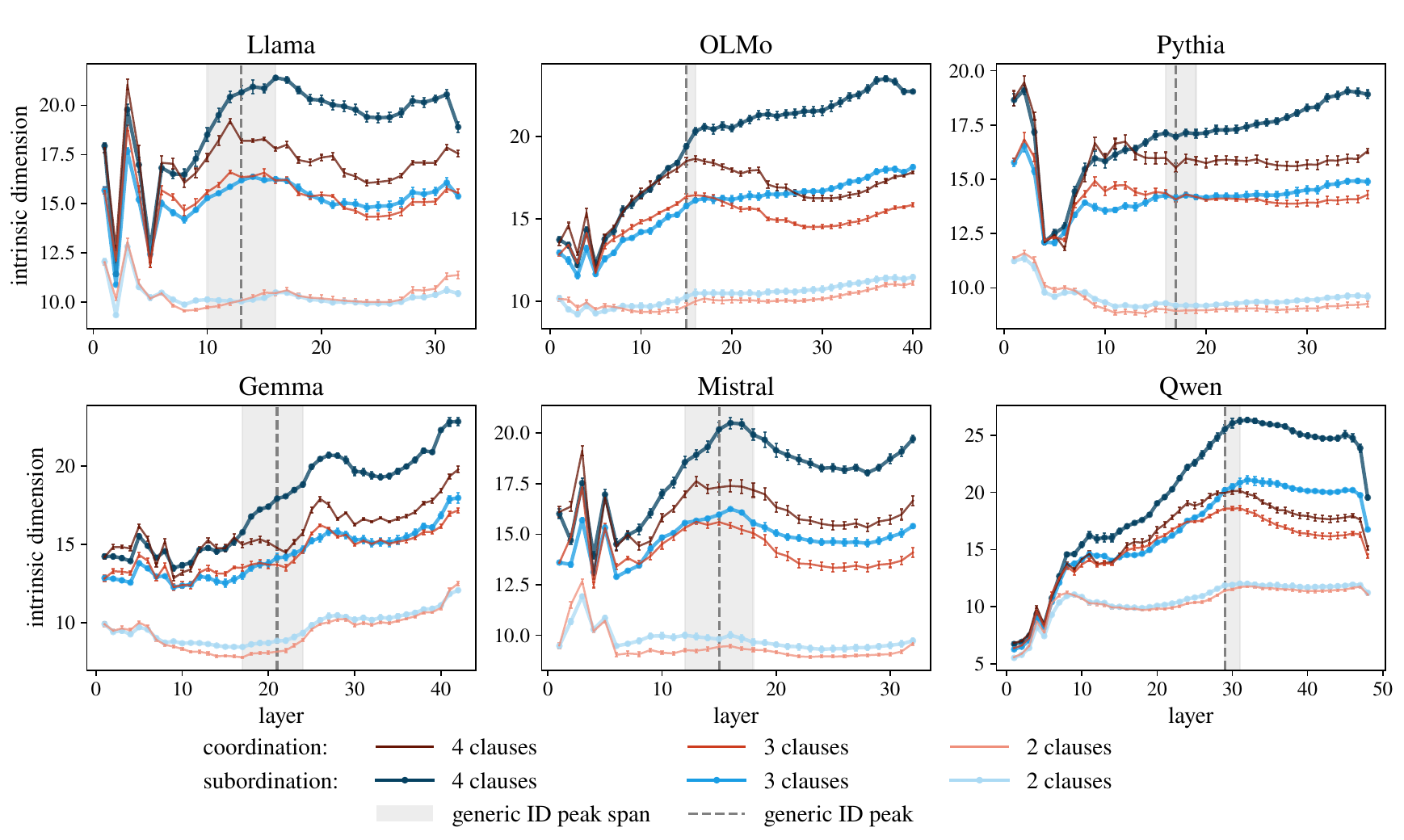}
    \caption{ID profiles through LLM layers (means and error bars across 5 partitions) for coordination/subordination sets featuring different numbers of clauses. Vertical dashed line marks maximum ID on generic sequences, and shaded area the corresponding span.}
    \label{fig:length_coord_subord}
\end{figure*}

In the main text, we present results for coordinated/subordinated sentences that contain 4 clauses. Fig.~\ref{fig:length_coord_subord} compares these results to those obtained when using 3-clause or 2-clause sentences (e.g., respectively: ``Quinn is rejoicing and/that Mary is screaming and/that the driver is faltering'' and ``Quinn is rejoicing and/that the driver is faltering''). We observe first of all that IDs, unsurprisingly, are higher for longer sentences. More importantly, the distinction between subordinated and coordinated sentences is consistent and strong in all models for the longest 4-clause sentences. It is present for most models also for 3-clause sentences, although typically with weaker separation. It is absent of very moderate for the shortest 2-clause sentences. It thus seems that the structural differences between flatter coordinated sentences and more hierarchical subordinated ones becomes salient enough to leave a recognizable trace in the ID profiles only above a certain degree of nesting.

\begin{figure*}[t]
    \centering
    \includegraphics[width=\textwidth]{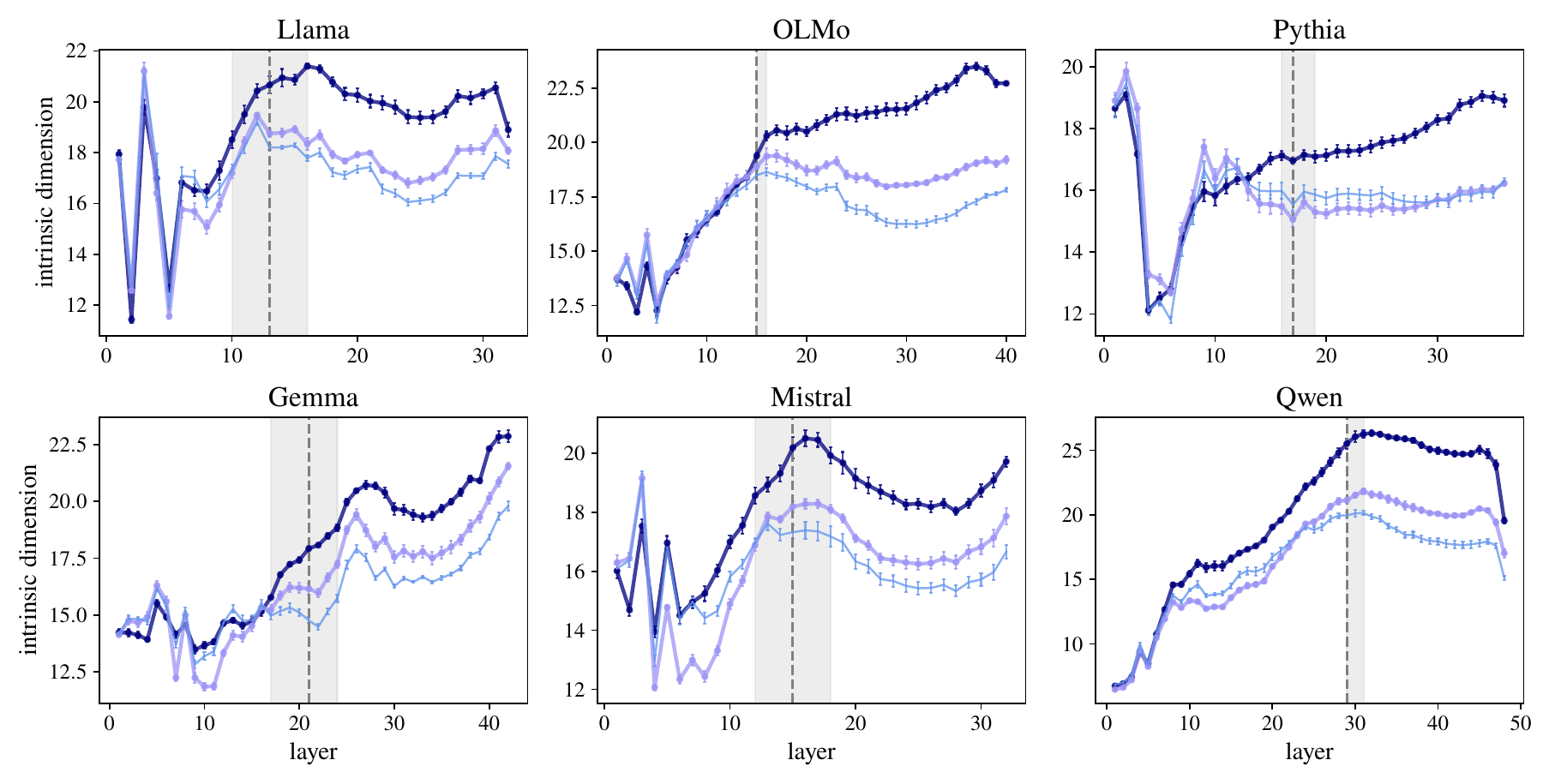}
    \includegraphics[width=\textwidth]{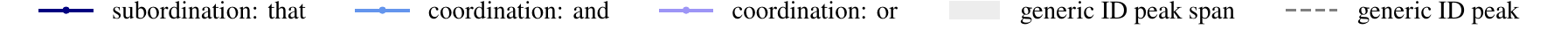}
    \caption{ID profiles through LLM layers (means and error bars across 5 partitions) for subordination vs.~coordination, including subordinated constructions using \textit{or}. Vertical dashed line marks
maximum ID on generic sequences, and shaded area the corresponding span.}
    \label{fig:or_and_that}
\end{figure*}

The stability of our results concerning the subordination/coordination contrast is confirmed by Fig.~\ref{fig:or_and_that}, where, for the 4-clause case, we include another coordination dataset, namely one where the sentences are coordinated by \textit{or} instead of \textit{and} (see Table \ref{tab:that-and-examples}). \textit{Or}-coordination displays lower ID than \textit{that-}subordination, with a profile that is qualitatively similar to that of \textit{and}-coordination. Interestingly, for all LLMs except Pythia, \textit{or}-coordination has higher ID than \textit{and}-coordination. We conjecture that this reflects higher semantic complexity, as \textit{or} introduces a set of alternatives over possible states of the world, rather than merely conjoining statements about the same state of the world.

\begin{figure*}
    \centering
    \includegraphics[width= \textwidth]{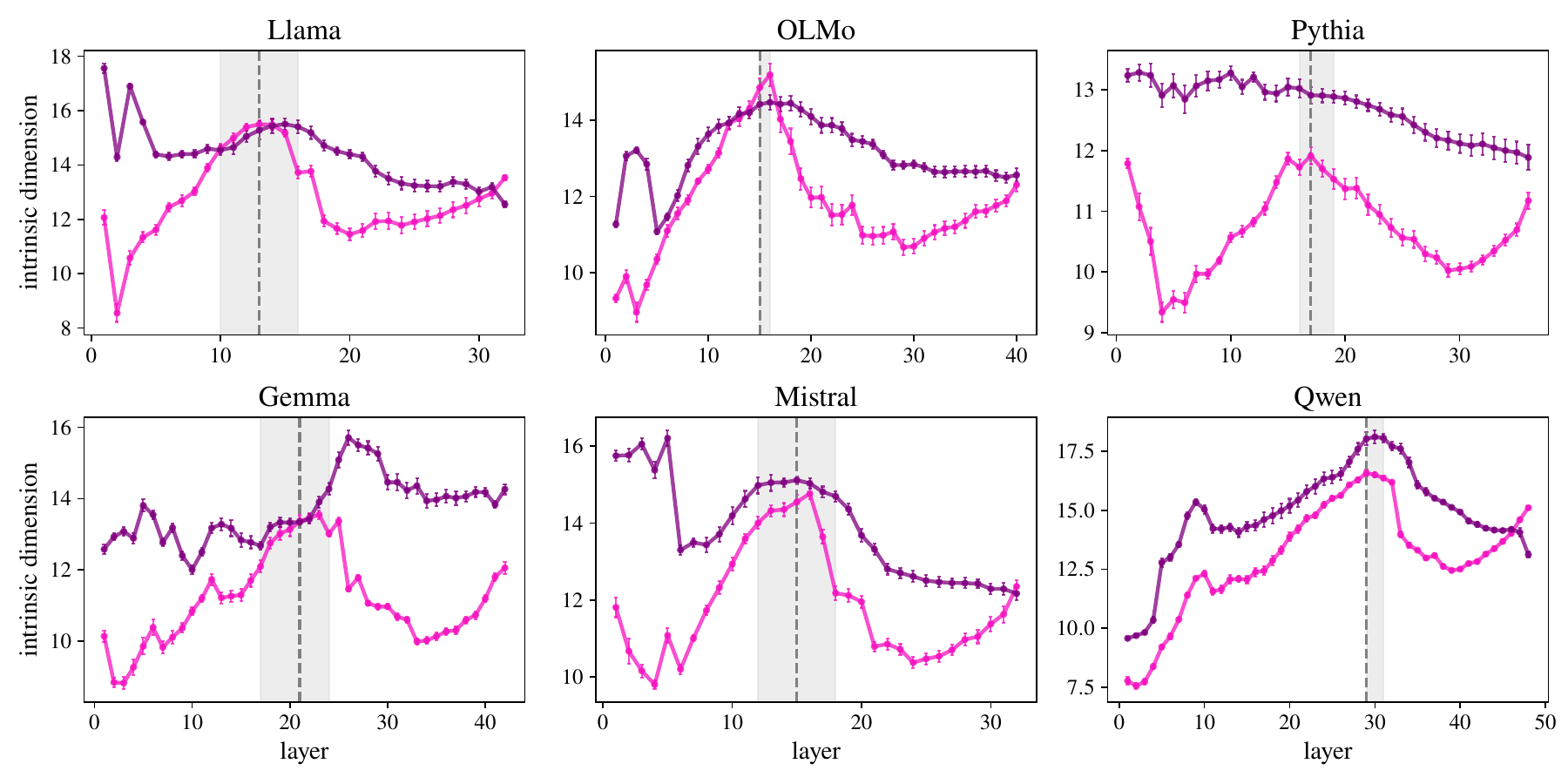}
    \includegraphics[width = \textwidth]{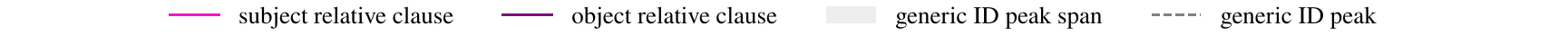} 
    \caption{ID profiles through LLM layers (means and error bars across 5 partitions) for the subject relative clause/object relative clause contrast.}
    \label{fig:src_orc_id}
\end{figure*}

\subsection{Intrinsic dimension for the subject vs.~object relative clause contrast}
\label{app:src_orc}
Besides the processing differences depending on the positions of relative clauses (right branching vs.~center embedding), we examined an additional condition based on embedded relative clauses. We compared the ID of sentences with object relative clauses (ORC), corresponding exactly to the sentences in the center-embedding condition, and sentences with subject relative clauses (SRC); Table~\ref{tab:center-right-examples} reports examples of both. Compared to ORCs, SRCs are supposedly less complex, as they involve shorter syntactic dependencies and are aligned with the subject-verb-object order. This is reflected in their easier parsing by humans~\citep{lau2021subject}. 

The ID patterns illustrated in Fig.~\ref{fig:src_orc_id} confirm this difference in complexity, showing consistently lower ID values across layers and models for SRCs. As in the right-branching/center-embedding contrast, the difference is often maximal in the pre-generic-ID layers. Interestingly, unlike right-branching structures, SRCs do not have high ID after the generic-ID peak. This supports our conjecture that the latter pattern in right-branching sentences is due to the fact that they end with an embedded clause (unlike center-embedded ORCs and SRCs). Another intriguing pattern is that SRCs (but not ORCs) have a clear peak coinciding with the generic one, so that the generic-ID-peak phase is actually the one where the difference between conditions is lowest.

\subsection{Intrinsic dimension profiles for the three relative clause attachment conditions}
\label{app:id_ambiguity}

\begin{figure*}[t]
    \includegraphics[width=\textwidth]{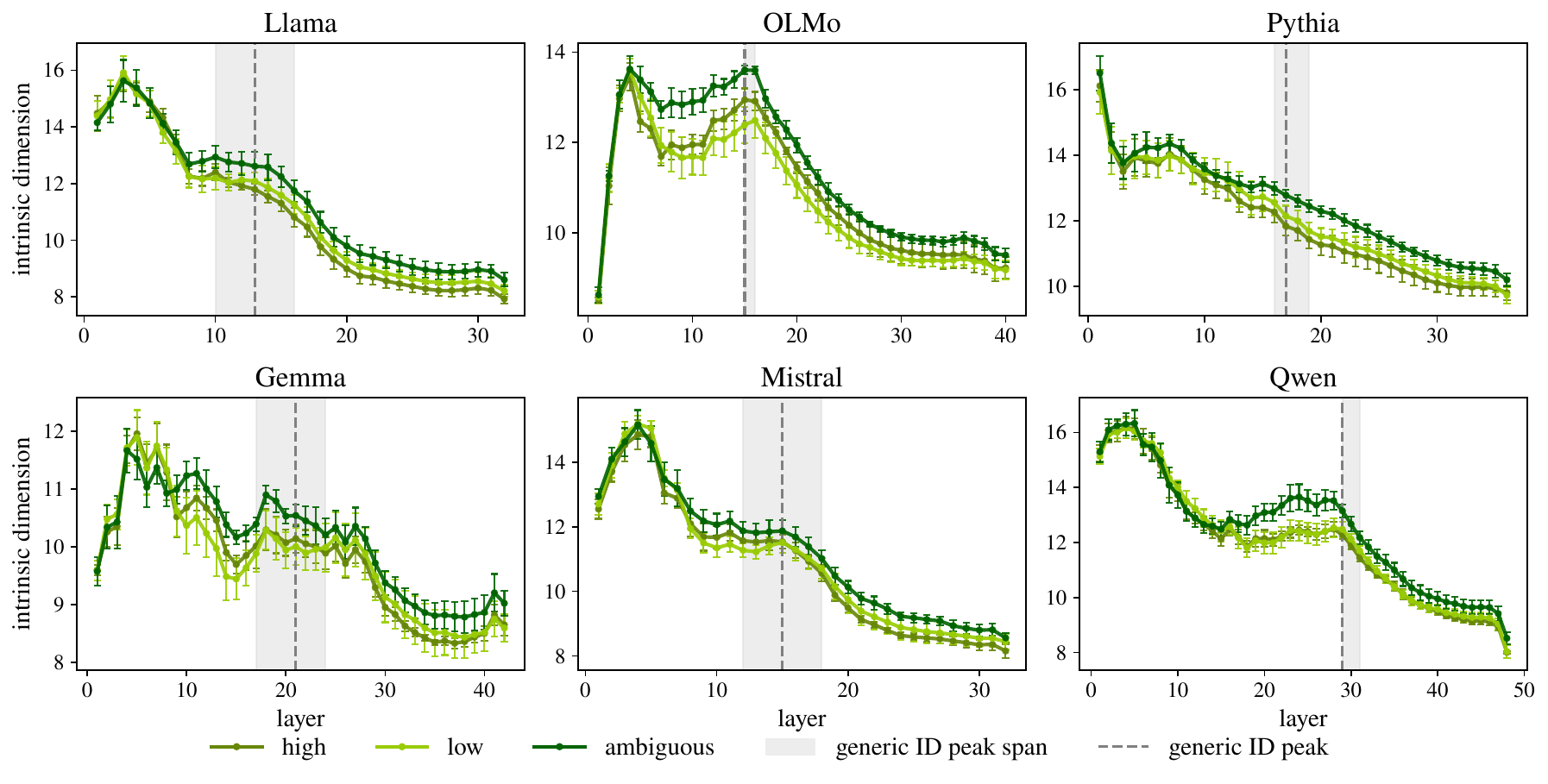}
    \caption{ID profiles through LLM layers (means and error bars across 5 partitions) for the high, low and ambiguous relative clause attachment conditions. Note that low attachment corresponds to the unambiguous condition shown in the last rows of figures \ref{fig:id_profiles_main} and \ref{fig:id_profiles_app}. Vertical dashed line marks maximum ID on generic sequences, and shaded area the corresponding span.}
    \label{fig:attachment_ambiguity_id_app}
\end{figure*}

In the main text  and App.~\ref{app:id_profiles}, we compared the ambiguous attachment condition to relative clauses with unambiguously low attachment. Fig.~\ref{fig:attachment_ambiguity_id_app} shows these two conditions together with a third, unambiguous \textit{high} attachment condition. We see that there are only small and non-systematic differences between the two unambiguous attachment conditions.

\subsection{Information Imbalance validation}
\label{app:ii_validation}

\begin{figure*}[h]
\centering
\includegraphics[width=\textwidth]{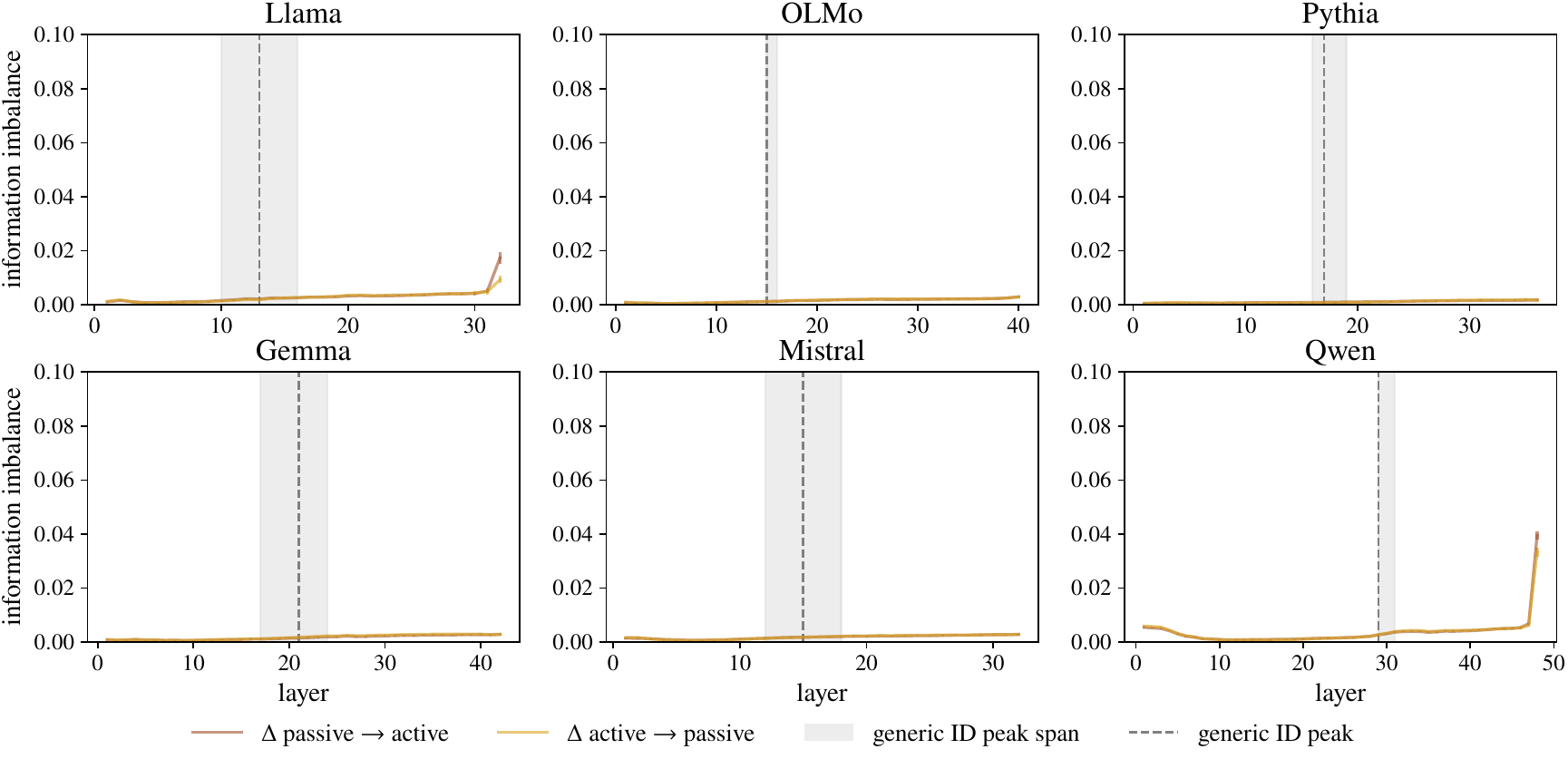}
\caption{Information Imbalance $\Delta$ between active and passive sentences for all models: means across 5 partitions with (present, but virtually invisible) error bars. Shaded area marks generic ID-peak span, with a vertical dashed line at the generic-ID maximum. Higher $\Delta$ means lower similarity.}
\label{fig:ii_profiles_active_passive}
\end{figure*}

To independently validate Information Imbalance as a measure of semantic information sharing, we created control contrasts between sentences expected to be nearly semantically equivalent, and between unrelated sentences. For the near-equivalent contrast, we chose the active/passive voice alternation. More precisely, we took the center-embedding dataset (third row of Table \ref{tab:experimental_manipulations}), and, for each sentence there, we created a variant with the relative clause turned into the passive voice. Thus, an example of the active/passive contrast is: ``The waiter that Bill humiliated was snickering'' vs.~``The waiter that was humiliated by Bill was snickering''.  The corresponding Information Imbalance profiles are shown in Fig.~\ref{fig:ii_profiles_active_passive}. As expected, Information Imbalance for these close paraphrases is consistently extremely low, approaching 0. Further, as expected for near-equivalent sentences, the Information Imbalance in the two directions is perfectly symmetrical. There is a very slight increase in Information Imbalance towards the final layers, which might be related to how the information structures associated to active vs.~passive might favor different continuations.

\begin{figure*}[h]
\centering
\includegraphics[width=\textwidth]{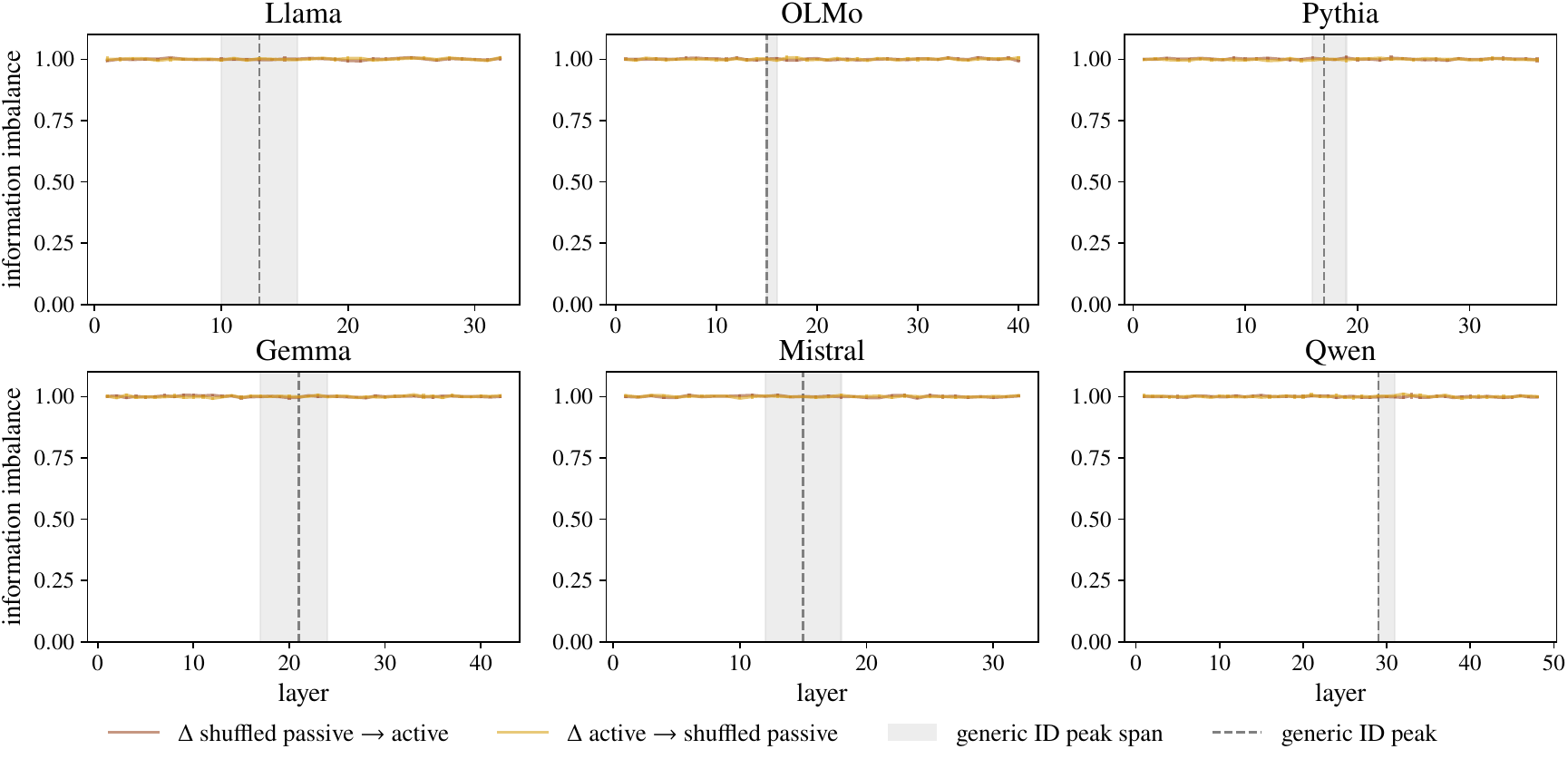}
\caption{Information Imbalance $\Delta$ between shuffled active and passive sentences for all models: means across 5 partitions with (barely visible) error bars. Shaded area marks generic ID-peak span, with a vertical dashed line at the generic-ID maximum. Higher $\Delta$ means lower similarity.}
\label{fig:ii_profiles_active_passive_shuffled}
\end{figure*}

We then used the same datasets, but now \textit{shuffling} the order of the passive sentences, so that each original active form would be paired with the passive of another active form. For example, ``The waiter that Bill humiliated was snickering'' might be paired with ``The potters that were advised by the politicians were waiting''. Fig.~\ref{fig:ii_profiles_active_passive_shuffled} shows that, for these unrelated sentences, Information Imbalance is consistently around 1, corresponding to no information sharing between the pairs \citep{Glielmo_Zeni_Cheng_Csányi_Laio_2022}.

The comparison between the active/passive and shuffled contrasts thus validates Information Imbalance as a way to capture semantic similarity between sentences. We point to \citet{acevedo2026a} for futher validation and theoretical comparisons between Information Imbalance and alternative information sharing measures.

\subsection{Information Imbalance profiles of the other models}
\label{app:ii_profiles}

\begin{figure*}[t]
\centering
\includegraphics[height=0.3\textheight]{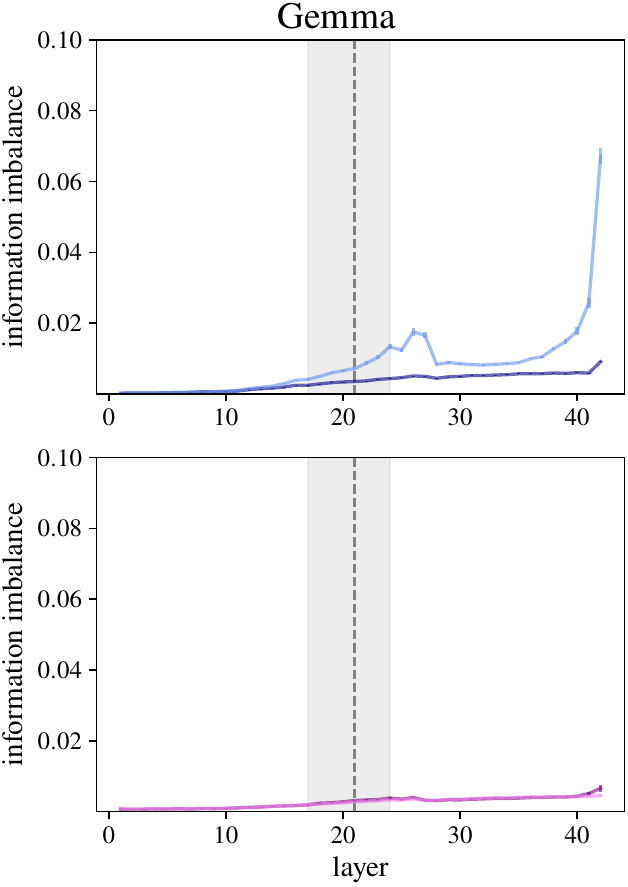}\hspace{2pt}
\includegraphics[height=0.3\textheight]{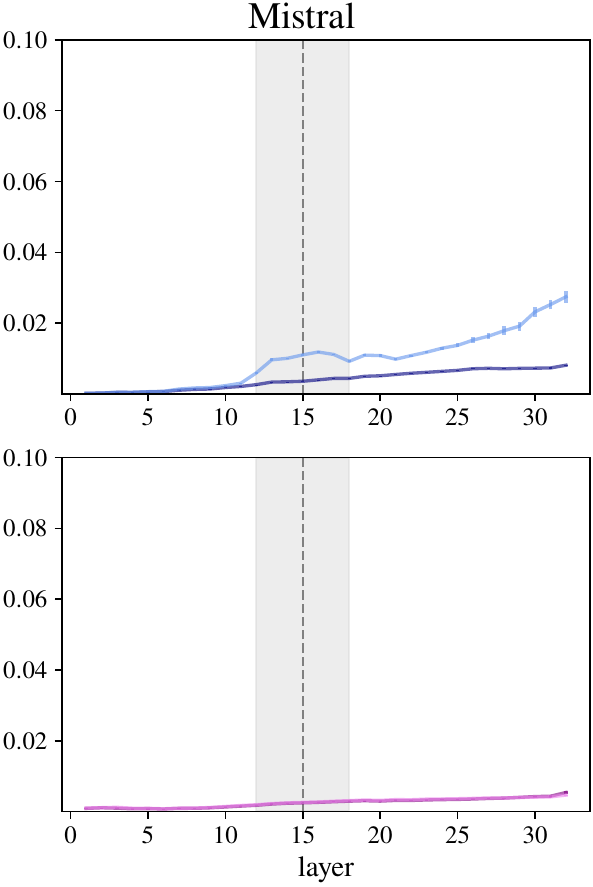}\hspace{2pt}
\includegraphics[height=0.3\textheight]{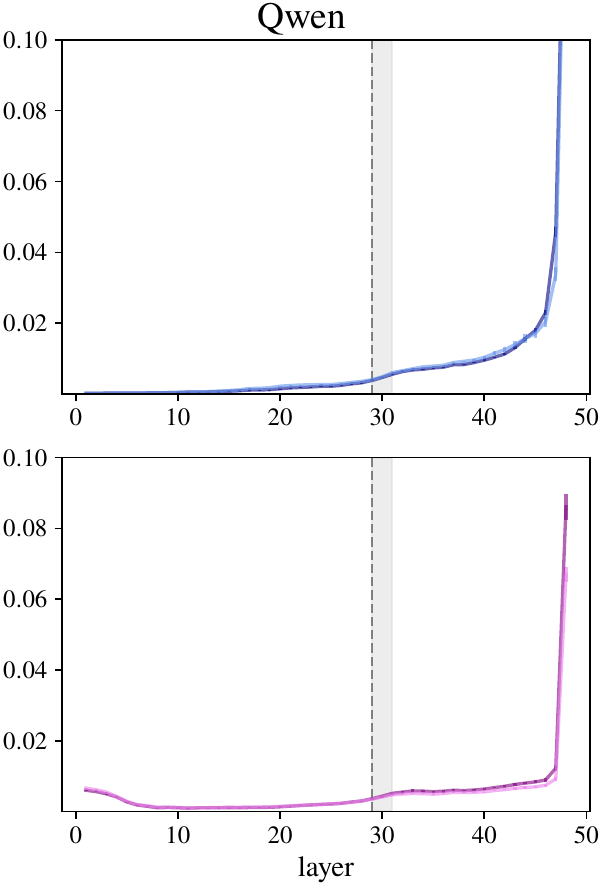}
\includegraphics[width=0.88\textwidth]{figures_v4/legend_ii.pdf}
\includegraphics[width=0.88\textwidth]{figures_v4/legendID_peak.pdf}
\caption{Information Imbalance $\Delta$ between coordinated/subordinated sentences (top) and right-branching/center-embedding sentences (bottom): means across 5 partitions with error bars (often so small as to be invisible),  for the Gemma, Mistral and Qwen models. Shaded area marks generic ID-peak span, with a vertical dashed line at the generic-ID maximum. Higher $\Delta$ means lower similarity.}
\label{fig:ii_profiles_app}
\end{figure*}

Fig.~\ref{fig:ii_profiles_app} reports the Information Imbalance ($\Delta$) results for the 3 models not shown in the main text (Gemma, Mistral and Qwen), confirming the trends we saw there.

\subsection{Pruning profiles of the other models and contrasts}
\label{app:ablation_profiles}

Pruning results with the coordinated/subordinated datasets for the 3 models not shown in the main text (Gemma, Mistral and Qwen) are reported in Fig.~\ref{fig:ablation_coord_subord_appendix}. The different responses to pruning are more visible in early layers and a dip in accuracy  specifically affecting the coordinated set is found in correspondence to the ID peak in Mistral, similar to Llama, OLMo and Pythia. 

\begin{figure*}[tb]
    \centering
     \includegraphics[width=\textwidth]{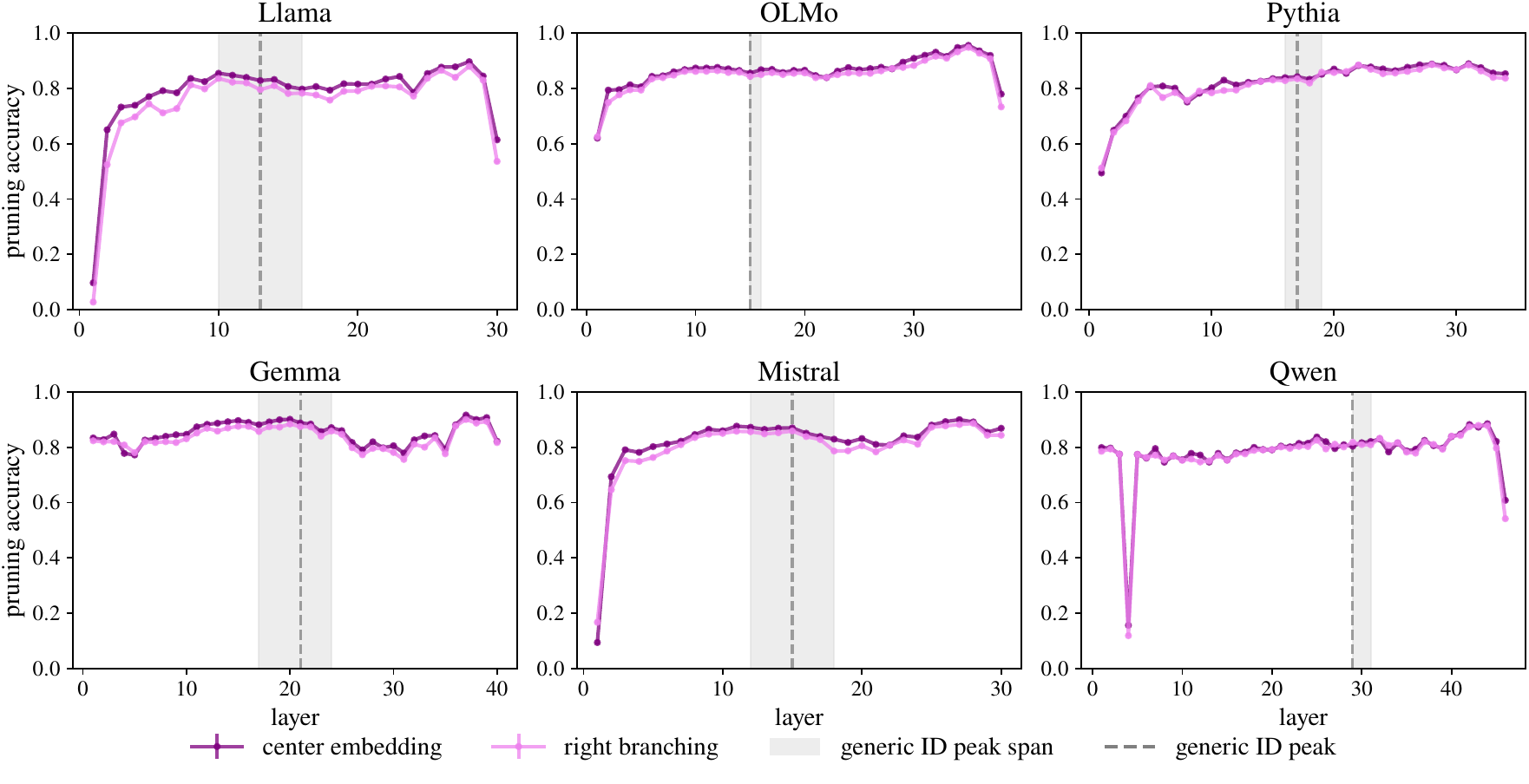}
    \caption{Pruning accuracy for right branching vs.~center embedding. Means and (hardly visible) standard errors over 5 partitions}
    \label{fig:ablation_center_right}
\end{figure*}

\begin{figure*}[tb]
    \centering
    \includegraphics[width=\textwidth]{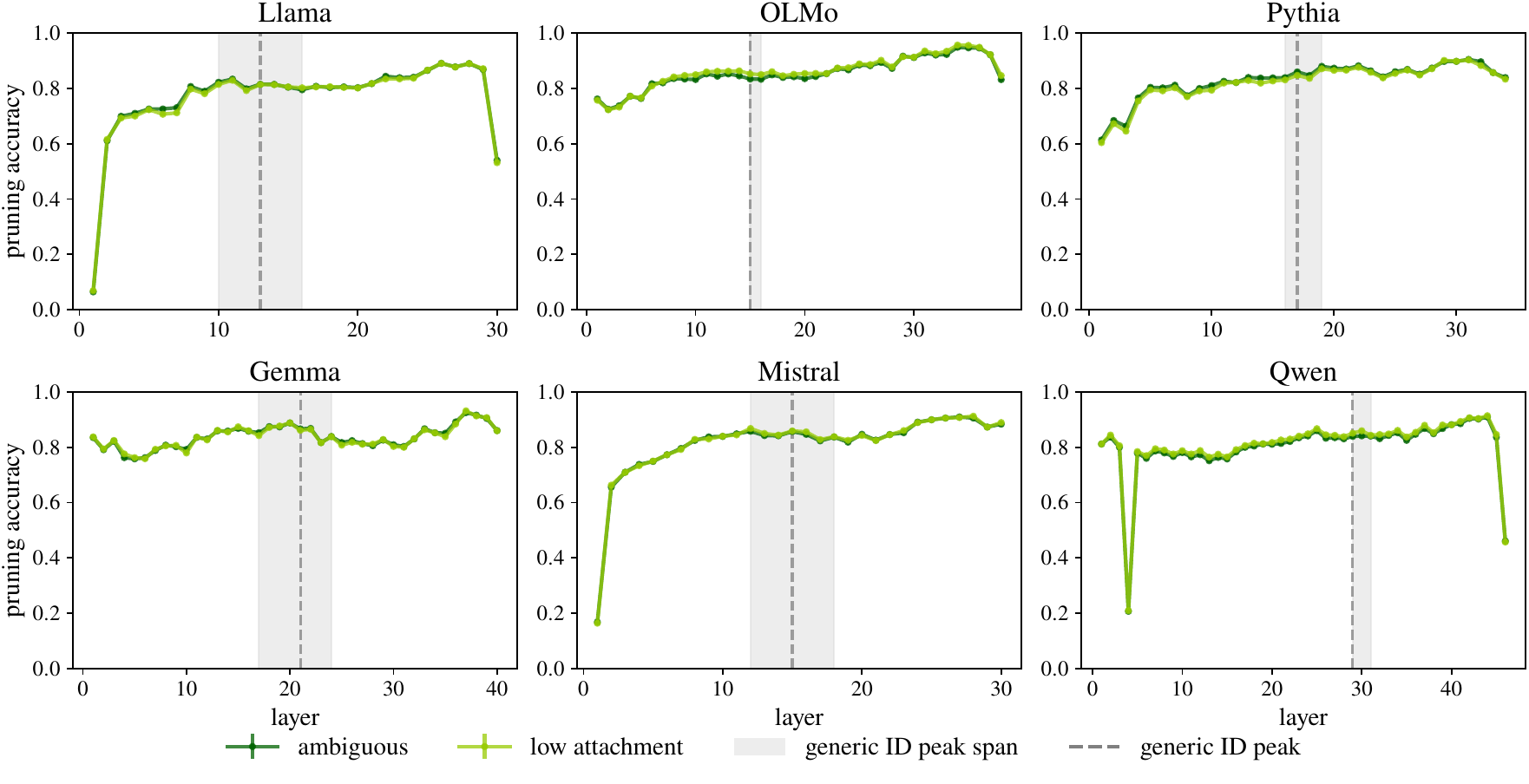}
    \caption{Pruning accuracy for the ambiguous vs.~unambiguous conditions. Means and (hardly visible) standard errors over 5 partitions.}
    \label{fig:ablation_ambiguity}
\end{figure*}

Pruning results for the other contrasts are reported in figures \ref{fig:ablation_center_right} and \ref{fig:ablation_ambiguity}, with  differences across conditions scarcely observable. 

\subsection{Pruning profiles with KL divergence}
\label{app:kl_ablation}

\begin{figure*}[tb]
\centering
 \includegraphics[width=\textwidth]{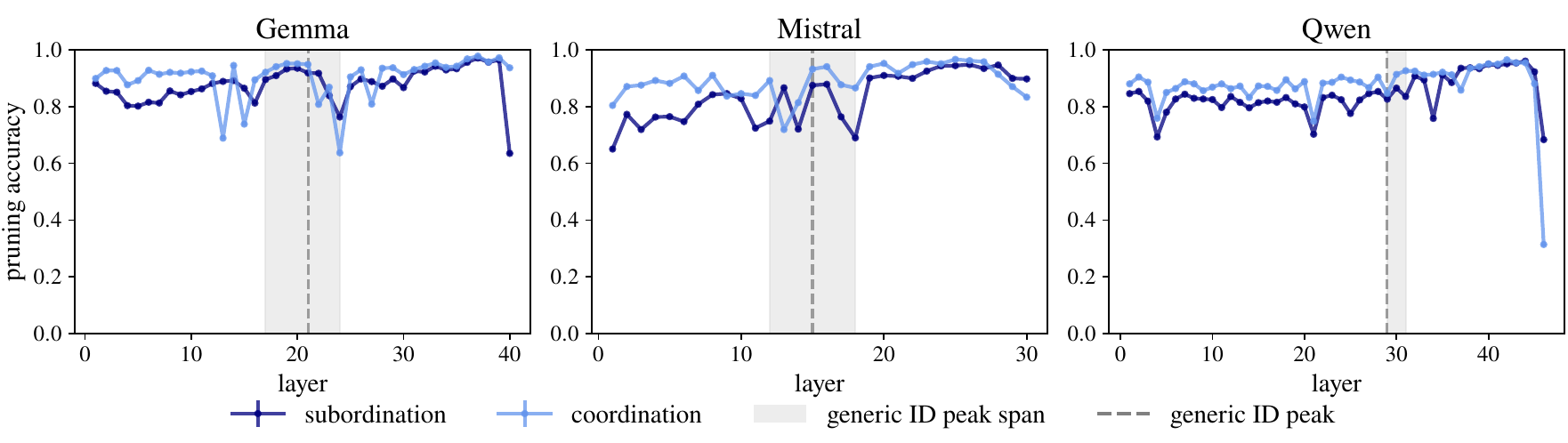}
\caption{Pruning accuracies comparing the coordination vs.~subordination conditions for Gemma, Mistral and Qwen. Means and (present but virtually invisible) standard error bars over 5 partitions.}
  \label{fig:ablation_coord_subord_appendix}
\end{figure*}

\begin{figure*}[tb]
\centering
 \includegraphics[width = \textwidth]{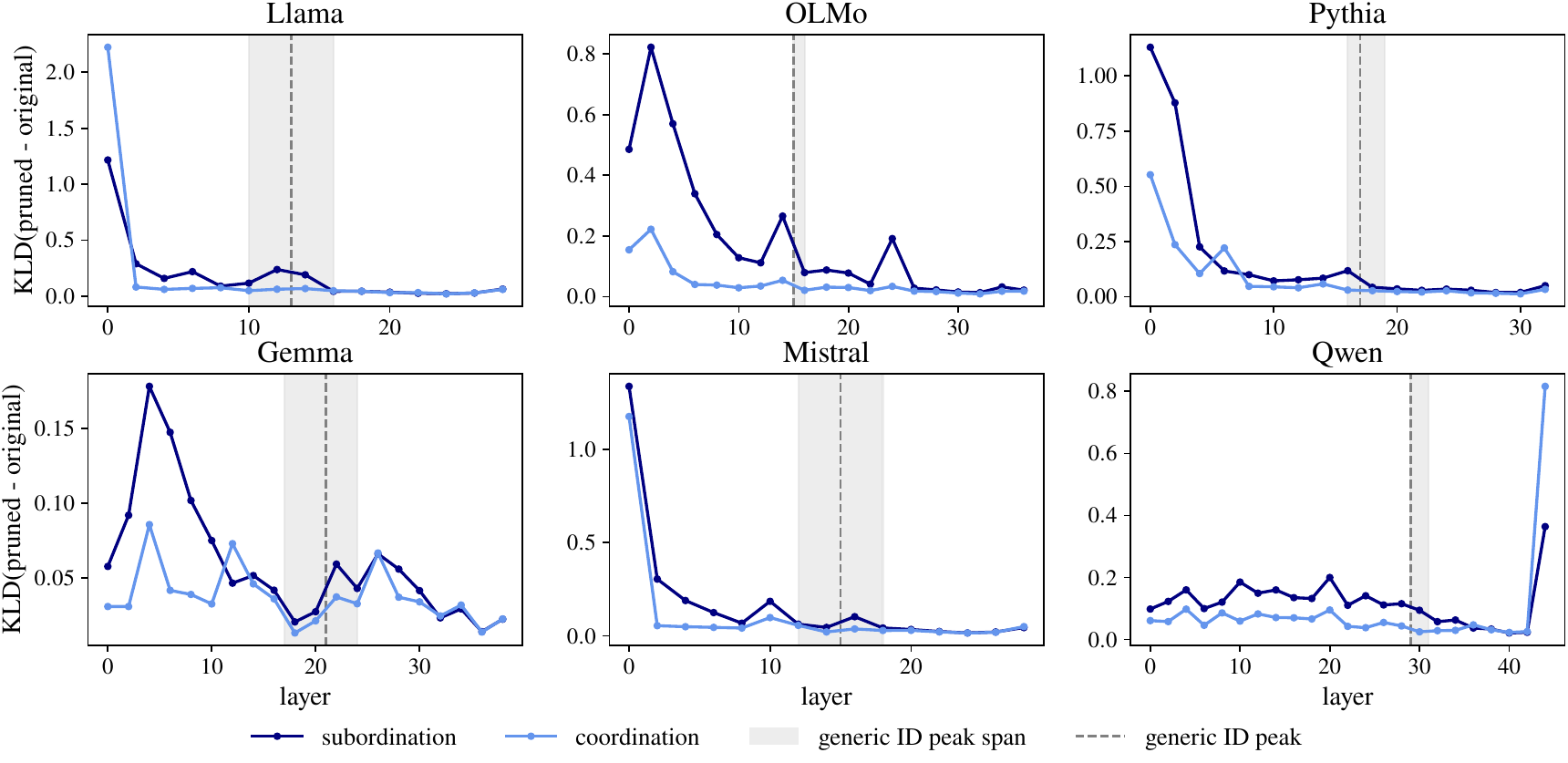}
\caption{Pruning effects measured through the KL divergence from the pruned to the original prediction, every two layers. For each model, the subordination (dark blue) and coordination (light blue) KL are shown against the pruned layer index. All curves show the mean over 5 random partitions, where one standard error is shaded (though very small, thus not visible in the plots).}
  \label{fig:kl_ablation_coord_subord_appendix}
\end{figure*}

Fig.~\ref{fig:kl_ablation_coord_subord_appendix} shows pruning profiles over layers for the crucial coordination/subordination contrast, where pruning impact is measured using KL divergence \citep{Kullback:Leibler:1951}, a continuous way to measure the difference between the next-token probability distribution before and after layer pruning. This analysis yields similar qualitative takeaways to those obtained through next-token prediction pruning accuracy. In particular, we observe upwards spikes (\textit{higher} KL means higher pruning impact) in the present plots that are aligned with the downwards dips (\textit{lower} accuracy means higher pruning impact) in the pruning accuracy plots of Figures \ref{fig:ablation_profiles_main} and \ref{fig:ablation_coord_subord_appendix}, in coincidence with the generic-ID peaks.


\end{document}